%% file: iclr2026_conference.tex
\documentclass{article} % For LaTeX2e
\usepackage{iclr2026_conference,times}

\input{math_commands.tex}

\usepackage{xcolor}
\usepackage{microtype}
\usepackage{hyperref}
\usepackage{url}
\usepackage{booktabs}
\definecolor{darkblue}{rgb}{0, 0, 0.5}
\hypersetup{colorlinks=true, citecolor=darkblue, linkcolor=darkblue, urlcolor=darkblue}
\usepackage{diagbox}
\usepackage{graphicx}
\usepackage{verbatim}
\usepackage{multirow}
\usepackage{algorithmic}
\usepackage{longtable}
\usepackage{array}
\usepackage{subfigure}
\usepackage{stfloats}
\usepackage{float}
\usepackage{balance}
\usepackage{multicol}
\usepackage{color}
\usepackage{epstopdf}
\usepackage{bm}
\usepackage{amsmath}
\usepackage{amssymb}
\usepackage{booktabs} % For formal tables
\usepackage{epsfig}
\usepackage{enumitem}
\usepackage{cleveref}
\usepackage{arydshln}
\usepackage{balance}
\usepackage{xspace}
\usepackage{enumitem}
\usepackage{soul}
\usepackage{mathtools}
\usepackage{bbm}
\usepackage[ruled,linesnumbered]{algorithm2e}
\usepackage{amsmath}
\usepackage{booktabs} 
\usepackage{makecell}
\usepackage{caption}
\usepackage{graphicx}
\usepackage{hyperref}
\usepackage{url}
\usepackage{graphicx}
\usepackage{tcolorbox}
\tcbuselibrary{breakable,skins}

\usepackage{array}
\usepackage{tabularx}
\usepackage{listings}
\usepackage{titlesec}

\usepackage{caption}
\usepackage{amsmath}
\usepackage{graphicx}
\usepackage{adjustbox}

\usepackage[utf8]{inputenc}

\newcommand{\vpara}[1]{\noindent \textbf{#1 }}

\usepackage{listings}
\usepackage{tabularx}

\definecolor{codegreen}{rgb}{0,0.6,0}
\definecolor{codegray}{rgb}{0.5,0.5,0.5}
\definecolor{codepurple}{rgb}{0.58,0,0.82}
\definecolor{codered}{rgb}{0.6,0,0}
\definecolor{backcolour}{rgb}{0.95,0.95,0.92}
\definecolor{backcolour2}{rgb}{0.95,0.92,0.95}

\lstdefinelanguage{SMTLIB}{
    keywords={
        set-logic, set-info, set-option, declare-sort, define-sort,
        declare-fun, define-fun, declare-const, define-const,
        assert, check-sat, get-model, get-value, get-proof,
        get-unsat-core, exit, push, pop, reset,
        and, or, not, ite, let, forall, exists, distinct,
        Int, Real, Bool, String, Array, BitVec,
        +, -, *, div, mod, abs, =, <, <=, >, >=,
        true, false,
        select, store,
        str.len, str.++, str.substr, str.contains, str.indexof,
        concat, extract, bvnot, bvand, bvor, bvxor
    },
    keywordstyle=\color{blue}\bfseries,
    keywords=[2]{
        sat, unsat, unknown, model
    },
    keywordstyle=[2]\color{magenta}\bfseries,
    comment=[l]{;},
    commentstyle=\color{codegreen}\itshape,
    string=[b]",
    stringstyle=\color{codepurple},
    basicstyle=\ttfamily\tiny,
    backgroundcolor=\color{backcolour},
    numbers=none,
    numberstyle=\tiny\color{codegray},
    breaklines=true,
    breakatwhitespace=false,
    keepspaces=true,
    showspaces=false,
    showstringspaces=false,
    showtabs=false,
    tabsize=2,
    frame=single,
    framerule=0.5pt,
    rulecolor=\color{gray!50}
}

\tcbset{sharp corners, boxrule=0.4pt}
\lstdefinestyle{solver}{
  language=SMTLIB,
  basicstyle=\ttfamily\scriptsize,
  numbers=none,
  columns=fullflexible,
  keepspaces=true,
  showstringspaces=false
}

\title{\method: 
Neuro-Symbolic Chain-of-Thought Validation via Logical Consistency Checks
}

\author{Yu Feng\textsuperscript{\rm \textdagger\thanks{Work done during internship at AWS}},
    Nathaniel Weir\textsuperscript{\rm $\spadesuit$},
    Kaj Bostrom\textsuperscript{\rm $\spadesuit$},  
    Sam Bayless\textsuperscript{\rm $\spadesuit$},\\
    \textbf{Darion Cassel\textsuperscript{\rm $\spadesuit$},
    Sapana Chaudhary\textsuperscript{\rm $\spadesuit$},
    Benjamin Kiesl-Reiter\textsuperscript{\rm $\spadesuit$}, 
    Huzefa Rangwala\textsuperscript{\rm $\spadesuit$}} \\
      $^{\text{\textdagger}}$University of Pennsylvania \quad 
  $^{\spadesuit}$Amazon Web Services\\
}

\newcommand{\method}{\textsc{VeriCoT}}
\newcommand{\methodlong}{\textsc{VeriCoT}: {Symbolic \textbf{Veri}fication for \textbf{C}hain-\textbf{o}f-\textbf{T}hought}}
\newif\ifcommentsoff % Toggle for disabling/enabling comments
\commentsofftrue  % Uncomment this line to disable comments
\newcommand{\authorcomment}[3]{\ifcommentsoff\unskip\space\ignorespaces\else\textcolor{#2}{[#1: #3]}\fi}

\newcommand{\sam}[1]{\authorcomment{Sam}{blue}{#1}}
\newcommand{\cut}[1]{}

\newcommand{\darion}[1]{\authorcomment{Darion}{olive}{#1}}
\newcommand{\ben}[1]{\authorcomment{Ben}{violet}{#1}}
\newcommand{\huzefa}[1]{\authorcomment{Huz}{cyan}{#1}}

\newcommand{\derivedset}{\mathcal{F}}
\newcommand{\premiseset}{\mathcal{P}}
\newcommand{\foconst}[1]{\mathit{#1}}
\newcommand{\fofunc}[1]{\foconst{#1}}
\newcommand{\fopred}[1]{\mathit{#1}}
\newcommand{\fovar}[1]{#1}
\newcommand{\newstatement}{F}
\newcommand{\nlstatement}{C}
\newcommand{\errors}{\textit{errors}}
\newcommand{\smtlib}{\texttt{SMT-LIB}\xspace}

\iclrfinalcopy % Uncomment for camera-ready version, but NOT for submission.
\begin{document}

\interfootnotelinepenalty=10000

% space hacking
\setlength{\abovedisplayskip}{6pt}
\setlength{\abovedisplayshortskip}{6pt}

% Set spacing after align environments  
% \setlength{\belowdisplayskip}{6pt}
% \setlength{\belowdisplayshortskip}{6pt}
% \setlength{\belowcaptionskip}{-10pt}
% \setlength{\abovecaptionskip}{5pt}
% \setlength{\belowdisplayskip}{5pt}
% \setlength{\belowdisplayshortskip}{5pt}
% \setlength{\textfloatsep}{30pt}
\setlength{\floatsep}{20pt}   
% \titlespacing*{\section}{0pt}{8pt}{4pt}

\setlength{\abovedisplayskip}{6pt}
\setlength{\abovedisplayshortskip}{6pt}
\setlength{\belowdisplayskip}{6pt}
\setlength{\belowdisplayshortskip}{6pt}
\setlength{\jot}{3pt}
\setlength{\textfloatsep}{5ex}
\titlespacing\section{0pt}{2pt plus 0pt minus 1pt}{1.5pt plus 0pt minus 1pt}
\titlespacing\subsection{0pt}{2pt plus 0pt minus 1pt}{1.5pt plus 0pt minus 1pt}

\maketitle

\input{0_abstract}
\input{1_introduction}
\input{2_method_revision}

\input{2a_method_modules}
\input{3_experiment}
\input{4_related_work}
\input{4b_limitations}
\input{5_conclusion}

% \clearpage
% \bibliography{iclr2026_conference,anthology}
\bibliography{iclr2026_conference.bbl}
\bibliographystyle{iclr2026_conference}

\clearpage
\appendix
\input{6_appendix}

\end{document}

%% file: math_commands.tex
\usepackage{amsmath,amsfonts,bm}

\def\eqref#1{equation~\ref{#1}}
\def\1{\bm{1}}

\DeclareMathAlphabet{\mathsfit}{\encodingdefault}{\sfdefault}{m}{sl}
\SetMathAlphabet{\mathsfit}{bold}{\encodingdefault}{\sfdefault}{bx}{n}

%% file: 0_abstract.tex
\begin{abstract}
LLMs can perform multi-step reasoning through Chain-of-Thought (CoT), 
but they cannot reliably verify their own logic. 
Even when they reach correct answers,
the underlying reasoning may be flawed, undermining trust in high-stakes scenarios.
To mitigate this issue, we introduce \method{}, a neuro-symbolic method that extracts and verifies formal logical arguments from CoT reasoning. 
\method{} formalizes each CoT reasoning step into first-order logic and identifies premises that ground the argument in source context, commonsense knowledge, or prior reasoning steps.
The symbolic representation enables automated solvers to verify logical validity
while the NL premises allow humans and systems to identify ungrounded or fallacious reasoning steps.
Experiments on the ProofWriter, LegalBench, and BioASQ datasets show \method{} effectively identifies flawed reasoning, and serves as a strong predictor of final answer correctness.
We also leverage \method{}'s verification signal for (1) inference-time self-reflection, (2) supervised fine-tuning (SFT) on \method{}-distilled datasets and (3) preference fine-tuning (PFT) with direct preference optimization (DPO) using verification-based pairwise rewards, further improving reasoning validity and accuracy.\ben{Here I tried to incorporate Nick's comment, which pointed out that previously we were trying to do too much here. I tried to make the progression 1-2-3 by going (1) signal is used for self-correction \emph{at inference time}, (2) signal is used directly for RFT \emph{at training time}, (3) signal is used to curate data \emph{before training} to then use it for training. That gives us a progression that goes further and further away from inference time.}
\end{abstract}

%% file: 1_introduction.tex
\section{Introduction}
\begin{figure}[t!]
\centering
\includegraphics[width=\textwidth, trim=0 0 85pt 0, clip]{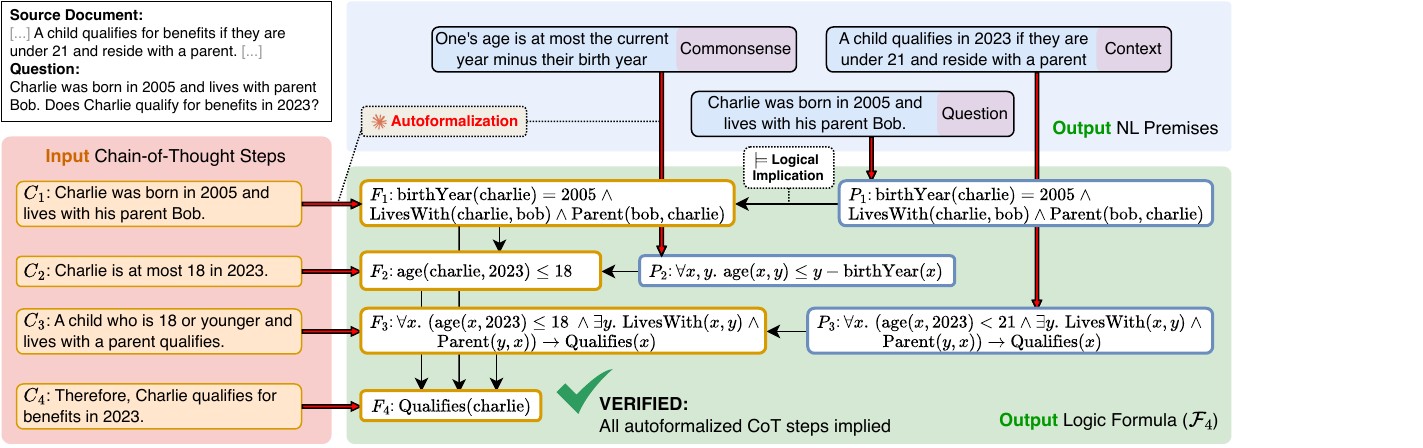}
\caption{\method{} verification of a \textcolor{orange}{Chain-of-Thought} for the SARA dataset~\citep{holzenberger2020datasetstatutoryreasoningtax}. 
Even if the final answer is correct, a CoT that contains an invalid step hurts user trust and raises questions of LLM faithfulness.
As shown in \S\ref{sec:example-verification}, \method{} autoformalizes each step of the CoT into symbolic logic, producing a \textcolor{codegreen}{formula} that ensures each one follows logically from a \textcolor{blue}{distilled list of NL premises}, each of which it \textcolor{codepurple}{annotates with their source type} (e.g. Commonsense or Context)
If the CoT cannot be represented this way, it is unverifiable.
}
\label{fig:motivating-example}
\end{figure}

Chain-of-Thought (CoT)~\citep{wei2022chain} in natural language (NL) has emerged as a powerful technique for improving large language model (LLM) performance on tasks that require reasoning. 
LLMs like DeepSeek-R1~\citep{deepseekR1} and OpenAI's o1~\citep{o1systemcard} demonstrate strong reasoning capabilities using CoT.
However, these models still frequently make logical errors in their reasoning chains, even when the final answer is correct~\citep{o1reasoningerrors}. 
For example, an LLM might generate the question answering (QA) CoT depicted in \autoref{fig:motivating-example}, which leads to a correct answer, but an intermediate step might be false (e.g., \textit{``Charlie is at most 15''} instead of \textit{``at most 18''}).
When LLMs generate logically invalid CoT steps, it hinders their trust and usefulness, regardless of whether the final answers are correct. This issue is particularly critical in high-stakes domains such as biomedical or legal reasoning, where users consider the accuracy of the reasoning path as important as that of the final answer.
This behavior can be attributed in part to an inherent limitation of LMs: they predict text without an explicit mechanism for verifying the logical validity of the resulting semantics~\citep{bender-koller-2020-climbing,ji-etal-2023-survey}

Prior work has attempted to mitigate this through self-refinement of the model's output based on diverse feedback sources, such as dynamic retrieval of facts from external knowledge bases~\citep{peng2023check}, a separate critic model~\citep{paul-etal-2024-refiner}, dynamic use of program execution~\citep{chen2024teaching, olausson2024is}, or symbolic checking~\citep{olausson-etal-2023-linc, quan-etal-2024-enhancing, lalwani2025auto, ye2023satlm, ling2023deductive, pan-etal-2023-logic}.
However, these approaches do not ensure the logical validity of the \emph{entirety} of an LLM's output.\ben{How about we consistently say LLM here instead of using the less common LLM? I get how we're trying to talk about language models more generally, and not just ``large'' ones, but is the distinction here helpful? And are there language models that can do chain of thought while totally not qualifying as ``large''?} The recently introduced Explanation-Refiner~\citep{quan-etal-2024-verification} makes progress towards this goal by performing iterative autoformalization and refinement of  NL explanations elicited from an LLM, guided by a theorem prover, for NLI task. However, there is yet to be a verification method that simultaneously (1) applies over LLM Chain-of-Thought steps, (2) formalizes the context grounding of every reasoning step, and (3) improves LLM reasoner capabilities with respect to logical validity in domains other than code/math.
To address these limitations, we introduce \methodlong{}, a neuro-symbolic algorithm that grounds and validates the logical consistency of Chain-of-Thought reasoning, identifying logical errors and making implicit premises explicit.

\method{} maintains a growing set of first-order-logic (FOL) premises inferred from NL context. 
It autoformalizes each CoT step into a first-order-logic formula, and uses a constraint solver to check whether it is  logically entailed by the premises and previously formalized steps.
If the step isn't entailed,
\method{} either autoformalizes a sufficient set of supporting premises from available context (question text, documents, or common sense) or identifies a reason why the step cannot be validated (e.g., because a premise could not be inferred from context, or because the step contradicts the inferred premises). 
Through this process, \method{} provides multi-faceted feedback. It  identifies \textbf{whether} a CoT can be represented in formal logic, \textbf{how} the CoT's steps are logically supported, and \textbf{what} underlying NL premises need to be accepted in order to accept the CoT's reasoning.
To our knowledge, this is the first neuro-symbolic validator of CoT traces in non-math/code domains.

Our evaluation demonstrates that \method{} can detect ungrounded or incorrect reasoning, and \method{}-validation is a strong predictor of final answer correctness, with validated CoT attaining higher precision than task-level accuracy (\S\ref{sec:effective}). 
Building on this verification capability, we leverage \method{} to actively enhance an LLM's reasoning (\S\ref{sec:experiments-applications}). 
First, we use \method{} for inference-time self-reflection, where its validity-oriented feedback prompts the model to self-correct, yielding an average 46\% relative improvement in CoT verification pass rate and consistent 41\% relative gains in producing accurate and verifiable task outcomes across multiple datasets.
Second, beyond inference-time correction, we build upon recent work using formal reasoning for RL verification~\citep{leang-etal-2025-theorem} and use \method{} to create a high-fidelity dataset of verified CoTs for Supervised Fine-tuning (SFT) and as a source of pairwise reward signals for preference fine-tuning with direct preference optimization (DPO). These fine-tuning strategies improve the model's ability to generate logically consistent CoTs by 18\% (relative) while matching or exceeding base task accuracy in less formal domains like biomedical and legal reasoning, since \method{} can supply stronger supervision signals especially when task labels are unavailable.

%% file: 2_method_revision.tex
\section{Neuro-Symbolic CoT Verification Algorithm}

Our goal is to validate the correctness of Chain-of-Thought reasoning by ensuring that individual reasoning steps can be translated into consistent first-order-logic formulas. 
We assume we are given a natural language \emph{context} (which includes a question, an optional conversation history, and a source document with information relevant to the question) and a \emph{CoT} consisting of NL steps $\nlstatement_1,\dots,\nlstatement_n$.  

As described in Alg.~\ref{alg:verification},  \method{} autoformalizes each step $\nlstatement_i$ into a first-order logic formula $\newstatement_i$, then uses a constraint solver to determine whether $\newstatement_i$ can be logically derived from our current knowledge~($\derivedset_{i-1}$) or requires additional premises.
We distinguish the following logical relationships between a new statement and our established knowledge:
\vspace{-1mm}
\begin{itemize}[leftmargin=20pt,itemsep=0em]
\item $\newstatement_i$ is \emph{entailed} by $\derivedset_{i-1}$, meaning it necessarily follows from what we already know ($\derivedset_{i-1} \models \newstatement_i$),
\item $\newstatement_i$ is \emph{contradicted} by $\derivedset_{i-1}$, meaning it is inconsistent with what we know ($\derivedset_{i-1} \models \neg \newstatement_i$),
\item $\newstatement_i$ is \emph{consistent} with, but not entailed by, what we already know ($\derivedset_{i-1} \nvDash \newstatement_i$, $\derivedset_{i-1} \nvDash \lnot \newstatement_i$).
\end{itemize}

\begin{algorithm}[t!]
\vspace{3pt}
\small
\begin{enumerate}[leftmargin=10pt]
\item \textbf{Initialize:} Set $\derivedset_0 = \emptyset$, $\premiseset_0 = \emptyset$, $\errors = \emptyset$.

\item \textbf{For each CoT step $\nlstatement_i$ ($i= 1, \ldots, n$):}
\begin{enumerate}[itemsep=3pt,leftmargin=*]
    \item \textbf{Autoformalization (\S\ref{sec:methdology:logical-translation}):} Formalize step $\nlstatement_i$ into logic formula $\newstatement_i$ . If $\nlstatement_i$ cannot be adequately expressed in our supported logic fragment or syntax errors persisted after multiple attempts: \\ \vspace{.5mm} 
    \quad $\errors\text{.add}(\langle i, \textbf{untranslatable}\rangle)$. Continue on to next CoT step $C_{i+1}$.

    \item \textbf{Consistency check:} If the formalized  $\newstatement_i$ contradicts our established knowledge ($\derivedset_{i-1} \models \neg \newstatement_i$): \\ \vspace{.5mm}
    \quad 
    $\errors\text{.add}(\langle i, \textbf{contradiction}\rangle)$. Continue on to next CoT step $C_{i+1}$.

    \item \textbf{Entailment check:} If $\newstatement_i$ is logically implied by our established knowledge ($\derivedset_{i-1} \models \newstatement_i$): \\ \vspace{.5mm} \quad set $\derivedset_i = \derivedset_{i-1} \cup \{\newstatement_i\}$ and $\premiseset_i = \premiseset_{i-1}$ and continue on to next step $C_{i+1}$. \\ \vspace{.5mm} Otherwise, try to generate supporting premises:
    
    \item \textbf{Premise generation (\S\ref{sec:methodology:knowledge-addition}):} Generate premise $P_i$ 
    to support $\newstatement_i$.\footnotemark \ Check if $P_i$ is consistent with established statements ($\derivedset_{i-1} \not\models \neg P_i$)   \\ \vspace{.5mm}
    \quad \emph{Optional}: \textbf{Use LLM-as-judge evaluation (\S\ref{sec:premise-evaluation})} to check that $P_i$ is attributable to source context. \\ \vspace{.5mm}
     
    \quad  If $P_i$ still fails to helps entail $\newstatement_i$ ($\derivedset_{i-1} \cup \{P_i\} \nvDash \newstatement_i$): 
$\errors\text{.add}(\langle i, \textbf{ungrounded}\rangle)$.  
    
    \vspace{.5mm} \quad set $\premiseset_i = \premiseset_{i-1} \cup \{P_i\}$ and $\derivedset_i = \derivedset_{i-1} \cup \{P_i\} \cup \{\newstatement_i\}$ and continue on to next step $C_{i+1}$.  \\ \vspace{.5mm}

\end{enumerate}

\item \textbf{Return:} Sets $\premiseset_n$, $\derivedset_n$, $\errors$.
\end{enumerate}
\caption{\method \ Overview}
\label{alg:verification}
\end{algorithm}

\method{} identifies a CoT as  \emph{valid} if it can infer a consistent set of logical premises from the NL context that are sufficient to entail every step of the CoT. If valid, \method{} returns:

\begin{enumerate}[leftmargin=*,itemsep=0em]
    \item A self-consistent set of FOL premises $\premiseset$, where each premise $P \in \premiseset$ represents the formalization of an NL source 
    (either from the provided context, or from common sense).
    \item For each NL step $C_i$, a FOL formula $\newstatement_i$ (the formalization of $C_i$) such that $\premiseset \vDash \newstatement_i$.
\end{enumerate}

If invalid, \method{} identifies error reasons for all steps that were not entailed. \darion{As per Nick's comment, we can expand on what the point of these errors is:} These reasons are used as a feedback signal for inference-time self-reflection or fine-tuning data distillation (\S\ref{sec:experiments-applications}):
\begin{enumerate}[leftmargin=*,itemsep=0em]
    \item \textbf{Ungrounded}:
    \method{} could not identify a sufficient, non-contradictory set of strengthening premises from context that would entail $\newstatement_i$ (e.g., $\derivedset_{i-1} \cup \{P_i\} \nvDash \newstatement_i$).
    \item \textbf{Contradiction}:  Existing statements  contradict $\newstatement_i$
    (e.g., $\derivedset_{i-1} \vDash \lnot \newstatement_i$).
    \item \textbf{Untranslatable}: $C_i$ could not be translated into the FOL subset supported by \method, or syntax errors persisted after multiple attempts.
\end{enumerate}
\label{Failure_reasons}

\footnotetext{$P_i$ can be a conjunction of multiple statements that are then evaluated individually by LLM-as-judge~(\S\ref{sec:premise-evaluation}).} 

Logic formulas are encoded in \smtlib~\citep{BarFT-SMTLIB}, which supports a fragment of first-order logic with theories including linear arithmetic, uninterpreted functions, and quantifiers. We use the SMT solver Z3~\citep{deMoura2008Z3AE} to perform the logical consistency and entailment checks described above.
To build intuition, we first walk through a simplified example that demonstrates our approach on the CoT depicted in \autoref{fig:motivating-example}. 

\subsection{Example of CoT Verification}
\label{sec:example-verification}

\vpara{Step 1: Infer premises from the user's question.} 
For the first step, ``\emph{Charlie was born in 2005 and lives with his parent Bob},'' \method{} autoformalizes it as follows (see \S\ref{sec:methdology:logical-translation}):
\begin{align*}
F_1:=  
\fofunc{birthYear}(\foconst{charlie}) = 2005 \land \fopred{LivesWith}(\foconst{charlie}, \foconst{bob}) \land \fopred{Parent}(\foconst{bob}, \foconst{charlie})
\end{align*}
\noindent At this initial point, the set $\derivedset_0$ of previously established statements is empty, so $\derivedset_0 \nvDash F_1$. 
Since the current step's statement cannot be derived from $\derivedset_0$, we check if supporting premises can be inferred from the provided context. 
Using an LLM, \method{} is able to derive a supporting premise $P_1$ from the question (in this case, $P_1$ happens to be identical to $F_1$).
We add $P_1$ to our premise set $\premiseset_1$, confirm (via constraint solver) that $\premiseset_1 \land \derivedset_0 \vDash F_1$, and finally, add both (identical) statements to $\derivedset_1$.

\vpara{Step 2: Infer premises from common sense.} 
For the second step, ``\emph{Charlie is at most 18 years old in 2023},'' \method{} autoformalizes it as:
\begin{align*}
F_2:= \fofunc{age}(\foconst{charlie}, 2023) \leq 18
\end{align*}
This statement cannot be derived from $\derivedset_1$, which contains only information about Charlie's birth year and living situation, but no direct facts about his age. To bridge this gap, our premise generation approach (see \S\ref{sec:methodology:knowledge-addition}) identifies a commonsense assertion relating age to birth year: ``\emph{Someone's age is at most the current year minus their birth year}'', autoformalized as:
\begin{align*}
P_2:= \forall \fovar{x}, \fovar{y}. \, \fopred{age}(\fovar{x}, \fovar{y}) \leq \fovar{y} - \fopred{birthYear}(\fovar{x})
\end{align*}
 We can derive $\fofunc{age}(\foconst{charlie}, 2023) \leq 18$ from the combination of $\derivedset_1$ and this new premise, so we obtain $\premiseset_2$ by adding $P_2$ to $\premiseset_1$, and we set $\derivedset_2 = \derivedset_1 \cup \{P_2, \newstatement_2\}$. Note that we reused the existing constant $\foconst{charlie}$ and the predicate $\fofunc{birthYear}$ from our established vocabulary, introducing only the new predicate $\fofunc{age}$ as needed. Also note that if the step had made an incorrect claim like ``\emph{Charlie is at most 15},'' our verification would report the step as inconsistent because $P_2$ contradicts it.

\vpara{Step 3: Infer premises from source document.} 
For the third step, ``\emph{A child who is 18 or younger and lives with a parent qualifies},'' we formalize it as:
\begin{align*}
F_3:= \forall \fovar{x}. \, (\fofunc{age}(\fovar{x}, 2023) \leq 18 \land \exists \fovar{y}. \, \fopred{LivesWith}(\fovar{x}, \fovar{y}) \land \fopred{Parent}(\fovar{y}, \fovar{x})) \rightarrow \fopred{Qualifies}(\fovar{x})
\end{align*}
This universal statement cannot be derived from $\derivedset_2$. 
However, the source document contains a stronger statement, which \method{} autoformalizes as a new premise, ``\emph{A child qualifies for benefits if they are under 21 and live with their parent}'', we formalize it as:
\begin{align*}
P_3:= \forall \fovar{x}. \, (\fofunc{age}(\fovar{x}, 2023) < 21 \land \exists \fovar{y}. \, \fopred{LivesWith}(\fovar{x}, \fovar{y}) \land \fopred{Parent}(\fovar{y}, \fovar{x})) \rightarrow \fopred{Qualifies}(\fovar{x})
\end{align*}

With $\premiseset_3$ strengthened by adding $P_3$ to $\premiseset_2$, we can derive $\premiseset_3 \land \derivedset_2 \models F_2$, and so we update $\derivedset_3 = \derivedset_2 \cup \{P_3, \newstatement_3\}$ accordingly.
Note that if this rule were not present in the source document or context, our verification would report the CoT step as ungrounded.

\vpara{Step 4: Conclusion.}
The last step, ``\emph{Therefore, Charlie qualifies for benefits in 2023},'' formalizes as:
\begin{align*}
F_4 := \fopred{Qualifies}(\foconst{charlie})
\end{align*}
This conclusion can now be derived from $\derivedset_3$ without requiring additional premises: we know Charlie is at most 18 (hence under 21), lives with his parent Bob, and the qualification rule applies to all children meeting these criteria. Had the step wrongly said ``\emph{Charlie does not qualify for benefits in 2023}'', we would have identified it as inconsistent.

%% file: 2a_method_modules.tex
Below we describe each modular component of \method{} in more detail.

\subsection{Autoformalization}
\label{sec:methdology:logical-translation}
Our autoformalization approach works in two stages, both using LLMs to translate NL into \smtlib. 
When attempting to translate a CoT step $C_i$ into its logical representation $\newstatement_i$, the first stage uses an LLM prompt that includes the previously produced logical vocabulary as context. 
The LLM is prompted to generate a structured, intermediate representation that combines \smtlib with metadata about which text in $C_i$ corresponds to which parts of the resulting $\newstatement_i$, using only the variables and types already present in the vocabulary.

The second stage extends the vocabulary if there are segments of the input text that the LLM deems relevant to the CoT's logical argument but cannot express with the existing vocabulary. 
In this stage, we use another LLM prompt to generate new \smtlib declarations (e.g., \texttt{declare-fun}, \texttt{declare-sort}). 
The new declarations are added to the vocabulary, and the first stage is attempted again. 
In our implementation, we allow this to repeat up to three times before giving up (in which case the current step is marked as untranslatable, or the current premise is discarded).

In the following example, when \method{} initially attempts to translate $C_2$ (``Charlie is at most 18 years old in 2023'') based on the available vocabulary, it first reports that the vocabulary is insufficient, then extends it accordingly to translate $C_2$ into $F_2$:
\begin{tcolorbox}[breakable, colback=white, colframe=white,  sharp corners, boxrule=0.4pt, title={}, left=1mm,           % Left margin
    right=1mm,          % Right margin  
    top=0mm,            % Top margin
    bottom=0mm,         % Bottom margin
    boxsep=0mm,         % Additional separation
    toptitle=0mm,       % Space above title
    bottomtitle=0mm     % Space below title
    ]
\noindent
\scriptsize
\begin{minipage}[t]{0.51\textwidth}

\textbf{Context:} Charlie was born in 2005 and lives with Bob.\\
\textbf{Document:} A child qualifies for benefits if they are under 21 and live with their parent.\\
\textbf{Question:} Does Charlie qualify for benefits in 2023?\\

\textbf{CoT Steps:}\\
\textcolor{codegreen}{$C_1$: ``Charlie was born in 2005 and lives with parent Bob"}\\
\textcolor{codered}{$C_2$: ``Charlie is at most 18 years old in 2023"}\\%[1.75mm]

\textbf{Previously Produced Vocabulary of $\derivedset_1$:} \\
\vspace{-4mm}
\begin{lstlisting}[language=SMTLIB, framerule=0.4pt, numbers=none]
; represents a person
(declare-sort Person)
; specific person Charlie
(declare-const charlie Person)
; specific person Bob
(declare-const bob Person)
; birth year of a person
(declare-fun birth_year (Person) Int)
; whether person lives with another person
(declare-fun lives_with (Person Person) Bool)
\end{lstlisting}
\end{minipage}%
\hfill%
\begin{minipage}[t]{0.44\textwidth}

\textbf{Initial (Failing) LLM Translation Output $\newstatement_2$:}\\[-4mm]
\begin{lstlisting}[language=SMTLIB, numbers=none, framerule=0.4pt]
; current year 2023 -- UNTRANSLATABLE
(assert false)
; charlie age <18 -- UNTRANSLATABLE
(assert false)
\end{lstlisting}

\vspace{2.4mm}
\textbf{New Declarations Added to Vocabulary:}  \\
\vspace{-4mm}
\begin{lstlisting}[language=SMTLIB, numbers=none, framerule=0.4pt]
; current year for calculation  
(declare-const current_year Int)
; age of a person in a given year
(declare-fun age_in_year (Person Int) Int)
\end{lstlisting}

\vspace{2.4mm}
\textbf{Successful LLM Translation Output $\newstatement_2$:}  \\
\vspace{-4mm}
\begin{lstlisting}[language=SMTLIB, numbers=none, framerule=0.4pt]
; current year 2023 
(= current_year 2023)
; charlie age <18 
(<= (age_in_year charlie current_year) 18)
\end{lstlisting}
\end{minipage}
\end{tcolorbox}

Successful autoformalization allows us to represent the step as $F_2$, but not necessarily ensure that it is entailed by the growing logical formula, which is the focus of \S\ref{sec:methodology:knowledge-addition}.

\subsection{Premise Generation}
\label{sec:methodology:knowledge-addition}

Chain-of-Thought steps aren't always directly implied by existing statements. Instead, they might rely on information from the context, like problem details (``Charlie is Bob's son'') or support documents (``a child qualifies for benefits if they are under 21''), or from common sense (``a father is a parent''). 
When a CoT step's formula $\newstatement_i$ is neither entailed nor contradicted by existing statements, \method{} prompts an LLM to identify supporting premises from the context or commonsense.

We attempt to build a premise formula $P_i$ that is sufficient to entail $\newstatement_i$ as follows: We first generate multiple noncontradictory candidate NL premises and translate each of them into a logic formula using the autoformalization process from \S\ref{sec:methdology:logical-translation}. We perform a round of premise regeneration if the natural language meaning of any new declarations from this step is not captured in the generated premises. For each candidate premise $p$, we then check whether it is consistent with the established statements (i.e., whether $\derivedset_{i-1} \land p$ is satisfiable) and keep only those that are. Finally, we conjoin all remaining candidate premises to form the final premise formula $P_i$.

\subsection{LLM-as-Judge Premise Evaluation}
\label{sec:premise-evaluation}
\method{} ensures that the CoT is FOL-representable and presents a list of inferred premises as one necessary basis for accepting the CoT's logic. However, it has not verified that whether the premises themselves might be accepted. It may have enumerated the premise ``the sky is purple'' as necessary for accepting the CoT, but this is generally not an acceptable statement on its own.
To provide increased assurance that the inferred premises are reliable, \method{} uses an LLM-as-Judge approach (LLMaj) to identify spurious premises after the premise generation process is complete. 
While the LLM generating premises in Alg.~\ref{alg:verification} can produce confabulations or omit relevant details, using LLM-as-Judge reduces the likelihood that these errors go undetected.

We evaluate all premises using the LLM-as-Judge approach. 
For premises inferred from source text, we provide judge LLMs with both the source document and the NL version of the premise, prompting them to evaluate whether the premise is attributable to the source text.
For premises inferred from common sense, we prompt judge LLMs to evaluate whether the premise is acceptable given the provided context and targeted reasoning step (omitting the attribution requirement).

%% file: 3_experiment.tex
\section{Experiments}
\label{sec:experiments}
\vpara{Evaluation Models.} We use Claude-3.5-Sonnet-V2~\citep{anthropic2024claude3_5_sonnet} through API calls as the executor of the proposed \method{}. We fine-tune Qwen2.5-7b-Instruct~\citep{qwen2.5}.

\subsection{Datasets}
We consider three datasets. \textbf{ProofWriter}~\citep{tafjord-etal-2021-proofwriter} contains small rulebases of facts and rules. Each rulebase has a set of questions which can either be proven true or false using proofs of various depths. \textbf{LegalBench}~\citep{guha2023legalbench} is a benchmark designed to evaluate the legal reasoning capabilities of LLMs crowdsourced from legal professional;
we use the \textbf{SARA}~\citep{holzenberger2020datasetstatutoryreasoningtax} subset from LegalBench, which evaluates statutory reasoning in tax law. \textbf{BioASQ}~\citep{Nentidis_2023, 450} is a series of annual open challenges focused on advancing biomedical semantic indexing and QA, specifically we use task b, which is biomedical QA with PubMed abstracts as context. 
Appendix Table~\ref{tab:datasets} shows data statistics.

\begin{table*}[t]
\centering
\begin{minipage}{0.56\textwidth}
\centering
\scalebox{0.8}{
\begin{tabular}{lcccc}
\toprule
\textbf{Method} & Pass Rate & Precision & VCAR & Task Acc \\
\midrule
\multicolumn{5}{c}{\textbf{ProofWriter}} \\
\midrule
ER             & 14.8 & 83.3 & 12.3 & 75.8 \\
DSB            & 10.0 & 96.1 & 9.5  & 74.8 \\
\method{}-NoPrem & 3.3  & 100  & 3.3  & 75.8 \\

\method{}      & 45.2 & 94.1 & \textbf{42.5} & 75.8 \\
\midrule
\multicolumn{5}{c}{\textbf{BioASQ}} \\
\midrule
ER             & 1.5  & 80.0 & 1.2 & 81.4 \\
DSB            & 5.9  & 72.2 & 4.2  & 75.7 \\
\method{}-NoPrem & 2.9  & 57.1 & 1.6  & 81.4 \\
\method{}      & 25.3 & 84.3 & \textbf{21.3} & 81.4 \\
\midrule
\multicolumn{5}{c}{\textbf{Legalbench-SARA}} \\
\midrule
ER             & 6.8 & 92.0 & 6.3 & 80.0 \\
DSB            & 4.8  & 94.1 & 4.5  & 77.7 \\
\method{}-NoPrem & 0.6  & 50.0 & 0.3  & 80.0 \\
\method{}      & 15.2 & 87.0 & \textbf{13.2} & 80.0 \\
\bottomrule
\end{tabular}}
\caption{Verification results across benchmarks. For each dataset, we report verification pass rate (\%), verifier precision (\%), verified correct answer rate (VCAR, \%), and task accuracy (\%, i.e., correct answer rate). \method{} consistently achieves the highest values across the main outcome metrics (pass rate and VCAR).}
\label{tab:vericot_initial}
\end{minipage}
\hfill
\begin{minipage}{0.4\textwidth}
\centering
\scalebox{0.85}{
\begin{tabular}{lc}
\toprule
\textbf{Category} & Score \\
\midrule
\multicolumn{2}{c}{\textbf{Legalbench-SARA}} \\
\midrule
Grounded contextual premise   & 93.5 \\
Acceptable commonsense premise & 83.9 \\
Necessary commonsense premise  & 77.0 \\
\midrule
\multicolumn{2}{c}{\textbf{BioASQ}} \\
\midrule
Grounded contextual premise   & 87.3 \\
Acceptable commonsense premise & 93.0 \\
Necessary commonsense premise  & 81.1 \\
\midrule
\multicolumn{2}{c}{\textbf{ProofWriter}} \\
\midrule
Grounded contextual premise   & 96.4 \\
Acceptable commonsense premise & 90.5 \\
Necessary commonsense premise  & 77.4 \\
\bottomrule
\end{tabular}}
\caption{Extended analysis of LLM-as-Judge premise evaluation breakdown. The generated premises of all types are highly acceptable under LLMaj.}
\label{tab:logical_soundness_onecol}
\end{minipage}
\end{table*}

\subsection{Baselines}
We compare \method{} against three baselines using Claude-3.5-Sonnet-V2. \textbf{Explanation-Refiner (ER)}~\citep{quan-etal-2024-verification}: an iterative auto-formalization and refinement approach guided by a theorem prover, originally proposed for NLI tasks. In our adaptation, we treat the CoT steps as the explanation, the question as the premise, and the model-generated answer as the hypothesis. \textbf{Direct SMT Baseline (DSB)}: A direct formalization approach that decomposes reasoning steps with designated types and generates \smtlib expressions simultaneously via few-shot prompting. For reasoning steps based on document, background, or commonsense premises, only a consistency check is applied; for inference steps, a entailment check is performed. \textbf{\method{} without Explicit Premise Generation (\method{}-NoPrem)}: A variant of our \method{} pipeline with generating intermediate premises turned off. Verification is performed using the same check protocol as in the Direct SMT Baseline. 

\subsection{\method{} as an Effective Verification Algorithm}
\label{sec:effective}

We first evaluate \method{} as a verification framework for CoTs, demonstrating that it achieves high verification performance by capturing a larger proportion of verifiable reasoning with strong precision and introduces an innovative mechanism for analyzing the premise grounding of reasoning steps.

\vpara{Direct evaluation of \method{}.} We evaluate verification methods using four complementary metrics: verification pass rate, verifier precision, verified correct answer rate (VCAR), and overall task accuracy. Pass rate measures how often CoTs are deemed verifiable. Precision captures the proportion of correct answers among verified CoTs, directly reflecting the reliability of \method{} decisions. VCAR combines these two by quantifying the overall fraction of CoTs that are both verified and correct. Finally, task accuracy corresponds to the final task-level correctness rate. As shown in Table~\ref{tab:vericot_initial}, \method{} achieves the highest pass rates across all benchmarks, leading to the strongest verified correct answer rate. Importantly, its precision is consistently high, exceeding the final task accuracy, demonstrating that \method{}-validated CoTs provide a more reliable signal of correctness than the raw CoTs. These results highlight \method{} as an effective and accurate verification algorithm that achieves broader coverage with better reliability.

\begin{figure*}[t]
\centering
\begin{minipage}{0.62\textwidth}
    \centering
    \includegraphics[width=.9\textwidth]{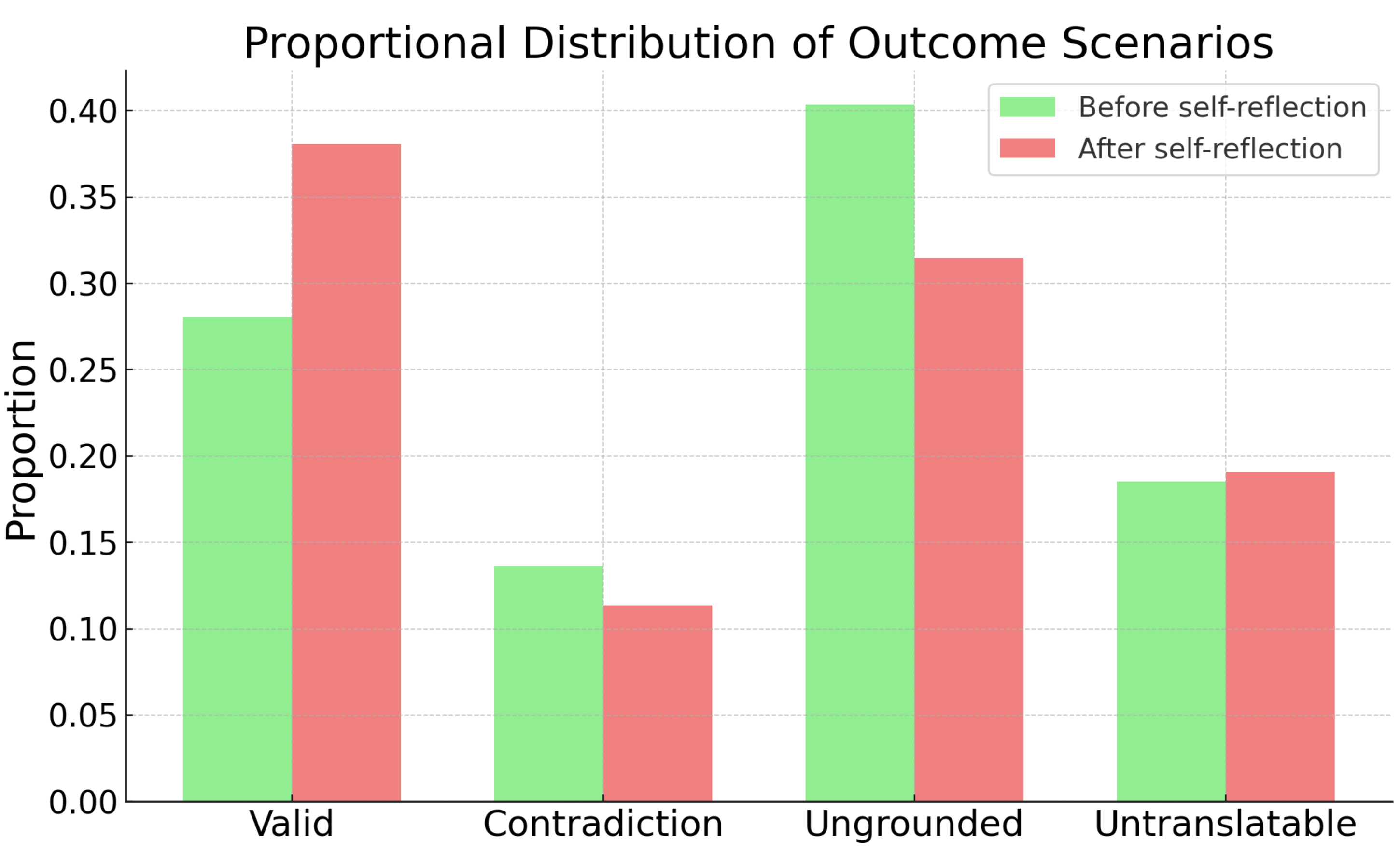}
\end{minipage}%
\begin{minipage}{0.34\textwidth}
    \caption{\label{fig:failure_modes} Proportional distribution of outcome scenarios before and after self-reflection (\S\ref{exp:inference}) under \method{}. Categories include successful verification (Valid) and failure cases: Contradiction, Ungrounded, Untranslatable as described in \S\ref{Failure_reasons}. Self-reflection significantly reduce errors.}
\end{minipage}
\end{figure*}

\vpara{Detailed analysis of instances that fail the \method{ }verification.}
We first conduct a quantitative analysis by presenting the proportional distribution of each possible outcome scenario under \method{} before and after self-reflection(introduced below in \S\ref{exp:inference}). Ungrounded error is the most prominent, highlighting that CoT often overgenerates assumptions, producing reasoning steps that appear plausible but unsupported. Self-reflection significantly increases the proportion of valid outcomes while reducing ungrounded and contradiction errors. The rate of translation-related errors remains largely unchanged. Additionally, we provide three illustrative examples in \S\ref{appendix:examples}, each corresponding to one type of failure case. For each example, we present the updated CoT along with its successful verification after self-reflection, as detailed in \S\ref{exp:inference}.

\vpara{Quantitative analysis of NL premise quality.}
We present the results of \method{} premise evaluation as described in \S\ref{sec:premise-evaluation}. For commonsense premises, we additionally evaluate whether the premise is necessary to support the corresponding reasoning step. The results in Table~\ref{tab:logical_soundness_onecol} confirm the high quality of premises identified by \method{}.  A key strength of \method{} is reliably locating reasoning steps in context or commonsense, making implicit premises explicit. For users to evaluate grounding, the challenge shifts from locating supporting premises to evaluating the quality of the premises that \method{} has already identified as shown in Appendix Tables~\ref{tab:example_proofwriter},~\ref{tab:example_bioasq} and ~\ref{tab:example_sara}.

\begin{table}[t]
\centering
\scalebox{0.80}{
\begin{tabular}{lccccccc}
\toprule
\textbf{Method} & Pass Rate & $\Delta$PR & Precision & $\Delta$Precision & VCAR & $\Delta$VCAR & Task Accuracy \\
\midrule
\multicolumn{8}{c}{\textbf{ProofWriter}} \\
\midrule
ER             & 21.5 & +6.7 & 80.2 & -3.1 & 17.2 & +4.9 & 75.8  \\
DSB            & 12.3 & +2.3 & 90.6 & -5.5 & 11.1 & +1.6 & 73.2 \\
\method{}-NoPrem & 30.7 & +27.4 & 88.8 & -11.2 & 27.3 & +24.0 & 72.4  \\
\method{}-Base      & 60.1 & +14.9 & 90.1 & -4.0 & 54.1 & +11.6 & \textbf{74.1}  \\
\method{}-w LLMaj   & \textbf{60.6} & +15.4 & 90.4 & -3.7 & \textbf{54.8} & +12.3 & 73.7 \\
\midrule
\multicolumn{8}{c}{\textbf{BioASQ}} \\
\midrule
ER             & 3.8 & +2.3 & 84.6 & +4.6 & 3.2 & +2.0 & 81.4  \\
DSB            & 8.5 & +2.6 & 76.9 & +4.7 & 6.5 & +2.3 & 78.8  \\
\method{}-NoPrem & 29.7 & +26.8 & 81.6 & +24.5 & 24.2 & +22.6 & 79.8  \\
\method{}-Base    & 33.4 & +8.1 & 86.3 & +2.0 & 28.8 & +7.5 & \textbf{82.0} \\
\method{}-w LLMaj   & \textbf{36.9} & +11.6 & 83.8 & -0.5 & \textbf{30.9} & +9.6 & 80.5  \\
\midrule
\multicolumn{8}{c}{\textbf{Legalbench-SARA}} \\
\midrule
ER             & 19.6 & +12.8 & 91.6 & -0.4 & 18.0 & +11.7 & 80.0  \\
DSB            & 9.0 & +5.2 & 93.8 & -0.3 & 8.4 & +3.9 & 80.2 \\
\method{}-NoPrem & 16.1 & +15.5 & 84.2 & +34.2 & 13.6 & +13.3 & 80.3  \\
\method{}-Base      & \textbf{25.2} & +10.0 & 85.5 & -1.5 & \textbf{21.6} & +8.4 & \textbf{81.4}\\
\method{}-w LLMaj    & 23.0 & +7.8 & 85.5 & -1.5 & 19.7 & +6.5 & 81.0 \\
\bottomrule
\end{tabular}}
\caption{Verification results after self-reflection. In addition to the metrics in Table~\ref{tab:vericot_initial}, we report absolute changes compared to pre-reflection ($\Delta$). \method{}-Base uses only verifier signals without LLMaj evaluation, while \method{}-w LLMaj leverages the full set of signals for self-reflection. \method{}(Base,w LLMaj) achieves the strongest improvements (VCAR and accuracy) across all benchmarks while maintaining precision at a similar or higher level compared to baselines.}
\label{tab:self_reflection}
\end{table}

\begin{table}[t]
\centering
\scalebox{0.85}{
\begin{tabular}{lcccc}
\toprule
\textbf{Method} & Pass Rate & Precision  & VCAR  & Task Accuracy  \\
\midrule
\multicolumn{5}{c}{\textbf{BioASQ}} \\
\midrule
Qwen2.5-7B-Instruct (direct) & 22.8 & 83.1 & 18.9 & 77.4 \\
+ SFT w Random Distilled CoTs & 22.9 & 85.7 & 19.6 & 78.5 \\
+ SFT w Verified CoTs & 22.9 & 85.5 & 19.6 & 79.7 \\
+ SFT w Verified CoTs + DPO & \textbf{26.8} & 83.5 & \textbf{22.3} & 79.4 \\
\midrule
\multicolumn{5}{c}{\textbf{ProofWriter}} \\
\midrule
Qwen2.5-7B-Instruct (direct) & 21.8 & 76.7 & 16.7 & 47.5 \\
+ SFT w Random Distilled CoTs & 24.6 & 71.4 & 17.6 & 47.3 \\
+ SFT w Verified CoTs & 23.2 & 81.5 & 18.9 & 51.1 \\
+ SFT w Verified CoTs + DPO & \textbf{27.8} & 82.8 & \textbf{23.0} & 51.8 \\
\bottomrule
\end{tabular}}
\caption{SFT and DPO results on BioASQ and ProofWriter. We report pass rate (\%), verifier precision (\%),  verified correct answer rate (VCAR, \%), and task accuracy (\%). Strategies include the base model (direct), fine-tuning with random distilled CoTs, verified CoTs, and DPO with verified CoTs.}
\label{tab:sft_results}
\end{table}

\subsection{Applications of \method{}’s Verification Signals}
\label{sec:experiments-applications}
The verification signals produced by \method{} provide structured feedback that improves CoT reasoning along three dimensions: user transparency, inference-time self-reflection, and fine-tuning. First, by auto-formalizing CoTs and making premises explicit, \method{} increases transparency, enabling users to directly and more effectively inspect the reasoning chain. Second, \method{} supports self-reflection at inference time by exposing granular errors in faulty CoTs, thereby providing actionable signals that help correct CoT. Third, the same structured signals can be incorporated into supervised fine-tuning and direct preference optimization (DPO)~\citep{rafailov2024directpreferenceoptimizationlanguage}, guiding models toward producing more verifiable CoTs and ultimately achieving higher task accuracy.

\vpara{Inference-time Self-reflection.}\label{exp:inference} We first evaluate whether  \method{} feedback can be used to guide models in producing better CoTs at inference time. Our process is as follows: if a CoT fails to pass the verification check, we prompt the model to self-correct its reasoning. We adopt two variants(Base, w LLMaJ): Base uses only verifier signals without LLMaJ premise evaluation, while \method{}-w LLMaJ leverages the full set of signals for self-reflection. Given all relevant information for every step, including the original reasoning steps, any added premises, the corresponding formalizations, errors, check results (with execution results and the model’s parameter assignment), and optional LLMaJ premise evaluation, the model is prompted to revise its CoT reasoning during inference. 
This generates an updated CoT, which is then re-evaluated.

Table~\ref{tab:self_reflection} reports verification pass rates, verifier precision, verified correct answer rates (VCAR), and final task accuracy after self-reflection, together with the corresponding improvements relative to pre-reflection values ($\Delta$PR, $\Delta$Precision, and $\Delta$VCAR). Across all benchmarks, \method{} with self-reflection yields consistent gains in both coverage ( average +12.3\%(absolute)/ +46.4\% (relative)) and verified correct answer rate ( average +9.5\%(absolute)/ +41.1\% (relative)), indicating that refinement encourages the generation of reasoning that is more frequently verifiable and leads to correct answers. Notably, \method{} achieves the highest absolute pass rates and verified correct answer rate, while maintaining a similar level of precision and slightly better final task accuracy. These results highlight the effectiveness of \method{}’s verification signals in guiding self-reflection. Note that w LLMaj with additional LLMaj evluation is only slightly better than Base. This is expected, as verifier results provide much more informative error signals.

\vpara{Supervised Fine-tuning.}
To harness the benefits of verification for model improvement, we build high-fidelity datasets consisting of verified, logically consistent CoT examples (pass both verifier checks and LLMaj evaluation) and apply them for supervised fine-tuning (SFT). We start from the Qwen2.5-7B-Instruct model and distill supervision signals from Claude-3.5-Sonnet-v2. Table~\ref{tab:sft_results} benchmarks different distillation strategies to evaluate the role of verification in data curation. Specifically, we compare (i) the baseline performance of Qwen2.5-7B-Instruct, (ii) SFT with randomly sampled distilled CoTs, (iii) SFT with distilled CoTs that pass \method{} verification. As shown in Table~\ref{tab:sft_results}, requiring CoTs to pass verification (ii vs. iii) yields an average 3\% improvement in final accuracy. This is because the verifier’s higher precision ensures that verified CoTs contain a greater proportion of correct answers than randomly selected ones. This highlights \method{}’s signal as particularly valuable, as it can distill better data when gold answers are unavailable.

\vpara{Preference Fine-tuning.}
The verification signals from each reasoning step can further be utilized as pairwise reward signals during preference fine-tuning (PFT) with DPO to enable more reliable CoT generation. We randomly resample CoTs for examples that initially pass \method{} verification, keep those that fail in the new attempt, and then construct chosen/rejected CoT pairs, where the chosen/rejected one passes/fails verification. 
When applying DPO on top of SFT , we observed two key improvements across all datasets as shown in Table~\ref{tab:sft_results}: the verification pass rate increased by 4.3\% (an 18.4\% relative improvement), while the verified correct answer rate improved by 3.4\% (a 17.7\% relative gain).
This demonstrates that preference fine-tuning on data generated by \method{} is especially helpful for generating verified CoTs because the pairwise reward signal provides a stronger supervision signal to distinguish correct reasoning traces from incorrect ones.

%% file: 4_related_work.tex
\section{Related Work}
\vpara{Solver-based Verification of Natural Language Reasoning.}
The most relevant work to ours is \cite{quan-etal-2024-verification}, where a neuro-symbolic pipeline is introduced to formalize natural-language explanations (for NLI) \huzefa{spell the acronym} into logical forms using LLMs, and then performs theorem-prover–based verification and refinement of those explanations. While \cite{quan-etal-2024-verification} focuses on verifying and refining explanations post hoc for NLI tasks, our framework not only translates and verifies each step of a CoT into first-order logic (\smtlib)\sam{If we are claiming FOL as a difference here, lets be explicit with how that is different: quan also formalizes to FOL, but restricts to the special form of event semantics - I dont know enough about event semantics to fully understand how that impacts expressiveness. A second, practical difference is that we target \smtlib rather than HOL/Isabelle, which allows for more compact formalization} but also grounds the reasoning chains in context and is tailored for multi-step reasoning beyond NLI explanations.\sam{Do we need to do a direct comparison to this work in our experiments section?} 
Additionally, work like \cite{ling2023deductive} and \cite{vacareanu2024generalpurposeverificationchain} uses LLMs to assess the logical consistency of each reasoning step, whereas our method adds formal solver checks, context grounding, and correction, yielding more objective verification than LLM-only pipelines.\sam{This is a claim that we should be showing empirically, eg in some kind of head-to-head experiment.}

\vpara{Solver-empowered Logical Reasoning in LLMs.}
Recent works have explored solver-assisted QA with LLMs~\citep{pan-etal-2023-logic, poesia2024certified,ye2023satlm}. Specifically, \cite{pan-etal-2023-logic} translates NL into logic and uses solvers with iterative repair; \cite{ye2023satlm} prompt LLMs to produce declarative constraints for SAT solving; \cite{poesia2024certified} constrain LLM outputs via symbolic state tracking. \cite{xu-etal-2024-faithful} propose a symbolic CoT that incrementally generates formal reasoning steps to encourage faithfulness to input context and logical constraints without a solver. We differ\sam{nit - differ from all the previous 4 works, or from Xu 2004 specifically?} by combining context-grounded \smtlib translation with step-wise verification and using solver feedback for on-the-fly reasoning correction.\sam{We should clarify this with a bit more detail. By on the fly correction, are you referring to inference time self reflection?}
The RL aspect of our paper is most related to \cite{leang-etal-2025-theorem}, who demonstrate the effectiveness of ``Theorem Prover-as-a-Judge'' feedback for learning to solve math problems.
The main difference is that their approach logically grounds all statements to a provided, symbolic context (Lean's mathlib), limiting it to math domains, whereas our approach logically grounds statements to logical premises inferred from NL context, allowing \method{} to be applied to arbitrary NL domains.

\vpara{Structured Explanation Generation for LLM Reasoning}
Our work relates to growing literature on surfacing the underlying premises and stepwise reasoning for task-specific conclusions. 
\method{} surfaces NL premises that, when formalized, provide a directed, tree-like logical basis for a symbolic conclusion; this is similar to the fully NL entailment tree structure~\citep{dalvi-etal-2021-explaining,bostrom-etal-2021-flexible} which uses NL textual entailment as the ``logical glue'' between each layer of inferences. Versions of entailment tree-generating algorithms such as Entailer~\citep{tafjord-etal-2022-entailer}, \textsc{Nellie}~\citep{weir-etal-2024-nellie}, and ADGV~\citep{sprague-etal-2022-natural} map from task answers to entailment trees rooted in underlying conjunctive premises. 
Maieutic Prompting~\citep{jung-etal-2022-maieutic} produces related structures rooted in underlying arguments for/against believing recursive inferences.

%% file: 4b_limitations.tex
\section{Limitations}
\label{sec:limitations}
As both autoformalization and premise inference rely on LLMs, it is possible that either step can be incorrect (either by mis-translating the CoT into an unrepresentative formula, or by introducing a premise that is not well-founded). 
A formalization may be incorrect simply because the LLM used made a mistake during translation, or because the text to be translated fundamentally cannot be represented in the subset of \smtlib that we support. As a result, while \method{} 
can prove that the formalization of the Chain-of-Thought necessarily follows from the inferred premises, it cannot guarantee that the NL CoT or the premises are correct.

%% file: 5_conclusion.tex
\section{Conclusion}
In this paper, we present a neuro-symbolic framework that validates Chain-of-Thought reasoning by autoformalizing each intermediate step into first-order logic and grounding it in formal premises inferred from NL context or commonsense knowledge. 
Across the ProofWriter, LegalBench, and BioASQ datasets, \method{} reliably detects ungrounded or incorrect reasoning in CoT traces, which we show is a strong predictor of final answer correctness.
We further show that these verification signals can guide models to self-correct during inference and through supervised fine-tuning and preference fine-tuning for better CoT reasoning.

%% file: 6_appendix.tex
\section{Appendix}
\subsection{Data Statistics}
\begin{table}[h]
\centering
\renewcommand{\arraystretch}{1.3} % add row spacing
\begin{tabularx}{\textwidth}{l c c X}
\toprule
\textbf{Dataset} & \textbf{Train Size} & \textbf{Test Size} & \textbf{Description} \\
\midrule
\textbf{ProofWriter} & 5000 & 400 & Randomly sampled from the OWA-Depth-5 and CWA-Depth-5 subsets. Contains only entailment/contradiction labels (no ``Unknown'' answers). Source: \url{https://huggingface.co/datasets/renma/ProofWriter}. \\
\textbf{BioASQ} & 5,049 & 340 & Restricted to Task 12b: Test Results of Phase B. \\
\textbf{LegalBench (SARA)} & - & 367 & 272 instances from SARA Entailment and 95 instances from SARA Numeric.\\
\bottomrule
\end{tabularx}
\caption{Dataset statistics and descriptions.}
\label{tab:datasets}
\end{table}

\subsection{Examples of failure cases under \method{}}
\label{appendix:examples}
We present three illustrative cases of different types of errors, each of which, after self-reflection, passes verification. Specifically, these include (1) an ungrounded example from ProofWriter as shown in Appendix Table~\ref{tab:example_proofwriter} (refer to full details in \ref{appnedix:example_proofwriter}): the model initially is over-claiming. During self-reflection, however, it is able to identify the overclaimed portion, rephrase it appropriately, and thus generate a verified CoT.

\begin{figure}[h]
\centering
\scriptsize
\adjustbox{width=\textwidth,center}{
\begin{tabular}{p{6cm}p{4cm}p{6cm}p{3cm}}
\toprule
\textbf{Question context} & \textbf{Output premise list} & \textbf{Output Logic (Omit less important ones)} & \textbf{Verification result} \\
\midrule
\textbf{Question:} 
Based on the above information, is the following statement true, false, or unknown? The mouse visits the rabbit

\textbf{Context (partially omitted):} 
... The mouse eats the tiger. The mouse is green. The rabbit does not chase the tiger. The rabbit eats the dog. ... If something visits the mouse then the mouse is big. If something eats the tiger then the tiger visits the mouse. ... If something is big then it visits the rabbit.

\textbf{Step-wise CoT:} 
\begin{enumerate}[wide=0pt,leftmargin=0pt,itemsep=0em]
\item The mouse has three direct properties: the mouse eats the tiger, the mouse is green, and there is a rule stating that if something visits the mouse then the mouse is big
 
\item Since the mouse eats the tiger and no other eating relationships exist except this explicitly stated, and there is a rule stating that if something eats the tiger then the tiger visits the mouse, we can conclude that the tiger visits the mouse

\item Since the tiger visits the mouse, and there is a rule stating that if something visits the mouse then the mouse is big, we can conclude that the mouse is big

\item Since the mouse is big, and there is a rule stating that if something is big then it visits the rabbit, we can conclude that the mouse visits the rabbit

\item The statement 'The mouse visits the rabbit' is true based on the logical chain of the mouse being big and the rule about big things visiting the rabbit
\end{enumerate}
& 
\textbf{Grounded in Context:}
\begin{enumerate}[wide=0pt,leftmargin=0pt,itemsep=0em]
\item The mouse is green
\item The mouse eats the tiger
\item If something visits the mouse then the mouse is big
\item If something eats the tiger then the tiger visits the mouse
\item If something is big then it visits the rabbit
\end{enumerate}

\textbf{Commonsense:}
\begin{enumerate}[wide=0pt,leftmargin=0pt,itemsep=0em]
\item The mouse and tiger are different animals
\item The rabbit and tiger are different animals
\item The rabbit and mouse are different animals
\end{enumerate}
& 
\begin{minipage}[t]{\linewidth}
\vspace{-2mm}
\begin{lstlisting}[language=SMTLIB, numbers=none,backgroundcolor={},frame=none, framesep=0pt,rulesep=0pt]
% [...]
(assert (animal_eats mouse tiger))

(assert (forall ((a Animal))
  (=> (animal_visits a mouse)
      (animal_is_big mouse))))

(assert (forall ((a Animal) (b Animal))
  (= (animal_eats a b)
     (and (= a mouse) (= b tiger)))))
      
(assert (=> (and (animal_eats mouse tiger)
                 (forall ((a Animal))
                   (=> (animal_eats a tiger)
                       (animal_visits tiger mouse))))
            (animal_visits tiger mouse)))

\end{lstlisting}
\end{minipage}
& 
Fail due to ungroundedness. In step 2, ``no
other eating relationships exist except this
explicitly stated'' with
\begin{minipage}[t]{\linewidth}
\vspace{-2mm}
\begin{lstlisting}[language=SMTLIB, numbers=none,backgroundcolor={},frame=none, framesep=0pt,rulesep=0pt]
(assert (forall ((a Animal) (b Animal))
  (= (animal_eats a b)
     (and (= a mouse) (= b tiger)))))
\end{lstlisting}
\end{minipage}
can not be grounded.
\\
\bottomrule
\end{tabular}
}
\caption{Highlights of an ungrounded example from Proofwriter before self-reflection.}
\label{tab:example_proofwriter}
\end{figure}

(2) a contradiction example from BioASQ as shown in Appendix Table~\ref{tab:example_bioasq} (refer to full details in \ref{appendix:exmaple_bioasq}): the reasoning contains a subtle logical flaw that is difficult to detect. Our VeriCoT framework successfully identifies this issue and prompts the model to explore an alternative reasoning path during self-reflection, ultimately resulting in  a verified CoT. 

\begin{figure}[h]
\centering
\scriptsize
\adjustbox{width=\textwidth,center}{
\begin{tabular}{p{4cm}p{4cm}p{6cm}p{3cm}}
\toprule
\textbf{Question context} & \textbf{Output premise list} & \textbf{Output Logic (Omit less important ones)} & \textbf{Verification result} \\
\midrule
\textbf{Question:} Can Connexin hemi channels be used for drug delivery? 

\textbf{Context omitted} 

\textbf{Step-wise CoT:} 
\begin{enumerate}[wide=0pt,leftmargin=0pt,itemsep=0em]
\item Connexin hemichannels are channels in cell membranes that can allow molecules to pass through and are part of gap junction channels.
\item Connexin hemichannels allow for the non-selective passage of molecules up to 1,000 Da.
\item The gap junction channel can transport macro polymer therapeutic molecules, which is evidence for drug delivery potential.
\item Therefore, Connexin hemi channels are also permeable to therapeutic macromolecules, makes them highly attractive targets for delivering drugs.
\end{enumerate}
& 
\textbf{Grounded in Context:}
\begin{enumerate}[wide=0pt,leftmargin=0pt,itemsep=0em]
\item Connexin hemichannels are channels in membranes that pass molecules.
\item Connexin hemichannels are part of gap junction.
\item Connexin hemichannels allow for the non-selective passage of molecules up to 1,000 Da.
\item The ability to deliver large polymeric therapeutic macromolecules via the gap junction route provides evidence for drug delivery potential.
\end{enumerate}

\textbf{Commonsense:}
\begin{enumerate}[wide=0pt,leftmargin=0pt,itemsep=0em]
\item Any macro polymeric therapeutic molecule has a molecular weight $> 1,000$ Da.
\end{enumerate}
& 
\begin{minipage}[t]{\linewidth}
\vspace{-2mm}
\begin{lstlisting}[language=SMTLIB, numbers=none,backgroundcolor={},frame=none, framesep=0pt,rulesep=0pt]
% [...]
(assert (forall ((m Molecule) (c Channel))
  (=> (and (IsConnexinHemichannel c)
           (CanEnterViaChannel m c))
      (<= (MolecularWeight m) 1000))))
(assert (forall ((m Molecule))
  (=> (IsMacroPolymerTherapeutic m)
      (> (MolecularWeight m) 1000))))
(assert (forall ((c Channel) (m Molecule))
  (=> (and (IsGapJunctionChannel c)
           (IsMacroPolymerTherapeutic m))
      (and (CanEnterViaChannel m c)
           (=> (CanEnterViaChannel m c)
               (HasDrugDeliveryPotential c)))
    )))
(assert (exists ((c Channel) (m Molecule))
  (and (IsConnexinHemichannel c)
       (IsMacroPolymerTherapeutic m)
       (CanEnterViaChannel m c))))
\end{lstlisting}
\end{minipage}
& 
Failed due to contradiction. 

In step 4, the axioms show macro-polymer molecules are $> 1000$ Da, but connexin hemichannels only permit molecules $\leq 1000$ Da. Thus, the same molecule would need to be both $\leq 1000$ and $> 1000$, which is impossible. \\
\bottomrule
\end{tabular}
}
\caption{Highlights of a contradiction example from BioASQ before self-reflection.}
\label{tab:example_bioasq}
\end{figure}

(3) an untranslatable example from LegalBench-SARA as shown in Appendix Table~\ref{tab:example_sara} (refer to full details in \ref{appendix:example_sara}): when the translation error is presented to the model for self-reflection, the model recognizes that the erroneous reasoning step is unnecessary for the final solution. By removing this step, the model produces a verified CoT.

\begin{figure}[h]
\centering
\scriptsize
\adjustbox{width=\textwidth,center}{
\begin{tabular}{p{6cm}p{4cm}p{5cm}p{3cm}}
\toprule
\textbf{Question context} & \textbf{Output premise list} & \textbf{Output Logic (Omit less important ones)} & \textbf{Verification result} \\
\midrule
\textbf{Question:} Determine whether the following statements are entailed under the statute.  Reply with either: Entailment, Contradiction.

\textbf{Context:}
Alice has paid \$3200 to Bob for agricultural labor done from Feb 1st, 2017 to Sep 2nd, 2017. Alice is an American employer.  
Statement: Section 3306(c)(1) applies to Alice employing Bob for the year 2017.

\textbf{CoT:} 
\begin{enumerate}[wide=0pt,leftmargin=0pt,itemsep=0em]
\item Section 3306(c)(1) of the statute addresses agricultural labor

\item The work performed by Bob for Alice is specifically agricultural labor

\item The basic scope of Section 3306(c)(1) covers all agricultural labor

\item While Alice is an American employer and there might be jurisdictional implications, the statute fragment does not provide information about geographical limitations

\item Bob's employment falls under the basic scope of Section 3306(c)(1)

\end{enumerate}
& 
\textbf{Grounded in Context:}
\begin{enumerate}[wide=0pt,leftmargin=0pt,itemsep=0em]
\item Section 3306(c)(1) applies to agricultural labor
\item Alice employed Bob to perform agricultural labor from February 1st to September 2nd, 2017, as evidenced by Alice's payment of \$3200 for this work
\end{enumerate}

& 
\begin{minipage}[t]{\linewidth}
\vspace{-2mm}
\begin{lstlisting}[language=SMTLIB, numbers=none,backgroundcolor={},frame=none, framesep=0pt,rulesep=0pt]
% [...]
(assert 
  (and 
    (employs alice bob)
    (performs_agricultural_labor bob) 
    (paid alice bob agricultural_labor 3200 20170201 20170902)))
(assert
  (forall ((p Person))
    (=> (performs_agricultural_labor p)
        (covered_by_section_3306c1 p))))
(assert  (covered_by_section_3306c1 bob))

\end{lstlisting}
\end{minipage}
& 
Fail due to untranslatability. In step 4, the phrase ``there might be jurisdictional implication'' cannot be rendered under SMI-LIB rules because it expresses a possibility rather than a definitive statement. \\
\bottomrule
\end{tabular}
}
\caption{Highlights of an untranslatable example from LegalBench-SARA before self-reflection.}
\label{tab:example_sara}
\end{figure}

\clearpage
\subsubsection{Ungrounded example from ProofWriter}
\label{appnedix:example_proofwriter}
\begin{tcolorbox}[breakable, colback=white, colframe=black, title={Ungrounded example from ProofWriter}]
\scriptsize
\textbf{Reference information:} The dog does not eat the rabbit. The mouse eats the tiger. The mouse is green. The rabbit does not chase the tiger. The rabbit eats the dog. The rabbit is young. The tiger is green. If something visits the mouse then the mouse is big. If something eats the tiger then the tiger visits the mouse. If the dog is young and the dog does not visit the mouse then the mouse does not visit the rabbit. If something is green and it chases the rabbit then the rabbit does not eat the mouse. If something is green then it visits the dog. If something visits the rabbit and the rabbit is young then it is round. If something is round and it visits the dog then it chases the tiger. If something is big then it visits the rabbit.

\vspace{2mm}
\textbf{Question:} Based on the above information, is the following statement true, false, or unknown? The mouse visits the rabbit.\\[1mm]
\textbf{Answer:} True

\vspace{2mm}
\textbf{Original CoT:} \\
Let me solve this step by step.\\[1mm]
1. First, let's identify what we know about the mouse:

* The mouse eats the tiger  

* The mouse is green  

* If something visits the mouse then the mouse is big

2. Since the mouse eats the tiger and and no other eating relationships exist except this explicitly stated, according to the rule:  

* "If something eats the tiger then the tiger visits the mouse"  

* Therefore, the tiger visits the mouse  

3. Since the tiger visits the mouse, according to the rule:

* "If something visits the mouse then the mouse is big"  

* Therefore, the mouse is big  

4. Now, since the mouse is big, according to the rule:  

* "If something is big then it visits the rabbit"  

* Therefore, the mouse visits the rabbit  

5. We can directly conclude that the mouse visits the rabbit based on these logical steps.  

Therefore, the statement "The mouse visits the rabbit" is TRUE.

\end{tcolorbox}

\method{} Output before self-reflection: 

\begin{tcolorbox}[breakable, colback=white, colframe=black, title={Decomposed CoT Steps}]
\scriptsize

\vspace{2mm}
1. The mouse has three direct properties: the mouse eats the tiger, the mouse is green, and there is a rule stating that if something visits the mouse then the mouse is big
 
2. Since the mouse eats the tiger and no other eating relationships exist except this explicitly stated, and there is a rule stating that if something eats the tiger then the tiger visits the mouse, we can conclude that the tiger visits the mouse

3. Since the tiger visits the mouse, and there is a rule stating that if something visits the mouse then the mouse is big, we can conclude that the mouse is big

4. Since the mouse is big, and there is a rule stating that if something is big then it visits the rabbit, we can conclude that the mouse visits the rabbit

5. The statement 'The mouse visits the rabbit' is true based on the logical chain of the mouse being big and the rule about big things visiting the rabbit
\end{tcolorbox}

\scriptsize

% ===== Step 1 =====
% ===== Step 1 =====
\begin{tcolorbox}[breakable, colback=white, colframe=black, title={Step 1}]
\scriptsize
\textbf{Step Context:} The mouse has three direct properties: the mouse eats the tiger, the mouse is green, and there is a rule stating that if something visits the mouse then the mouse is big.

\vspace{2mm}
% full-width inner box
\begin{tcolorbox}[breakable, colback=black!2, colframe=black]
\begin{lstlisting}[style=solver]
; =======================
; Schema
; =======================
; represents an animal entity
(declare-sort Animal)
; constant representing the mouse
(declare-const mouse Animal)
; constant representing the tiger
(declare-const tiger Animal)
; true if one animal eats another
(declare-fun animal_eats (Animal Animal) Bool)
; true if an animal is green
(declare-fun animal_is_green (Animal) Bool)
; true if an animal is big
(declare-fun animal_is_big (Animal) Bool)
; true if one animal visits another
(declare-fun animal_visits (Animal Animal) Bool)

; =======================
; Added premises
; =======================
; The mouse is green (document)
(assert (animal_is_green mouse))
; The mouse and tiger are different animals (commonsense)
(assert (not (= mouse tiger)))
; The mouse eats the tiger (document)
(assert (animal_eats mouse tiger))
; If something visits the mouse then the mouse is big (doccument)
(assert (forall ((a Animal))
  (=> (animal_visits a mouse)
      (animal_is_big mouse))))

; =======================
; Step logic
; =======================
; mouse eats tiger
; mouse is green
; if animal visits mouse then mouse is big
(assert (animal_eats mouse tiger))
(assert (animal_is_green mouse))
(assert (forall ((a Animal))
  (=> (animal_visits a mouse)
      (animal_is_big mouse))))
\end{lstlisting}
\end{tcolorbox}

\vspace{1mm}
\textbf{Evaluation:} \textcolor{green!40!black}{Pass}
\end{tcolorbox}

% ===== Step 2 =====
\begin{tcolorbox}[breakable, colback=white, colframe=black, title={Step 2}]
\scriptsize
\textbf{Step Context:} Since the mouse eats the tiger and no other eating relationships exist except this explicitly stated, and also there is a rule stating that if something eats the tiger then the tiger visits the mouse, we can conclude that the tiger visits the mouse.

\vspace{2mm}
\begin{tcolorbox}[breakable, colback=black!2, colframe=black]
\begin{lstlisting}[style=solver]
; =======================
; Schema
; =======================
; (none new)

; =======================
; Added premises
; =======================
; If something eats the tiger then the tiger visits the mouse (document)
(assert (forall ((a Animal))
  (=> (animal_eats a tiger)
      (animal_visits tiger mouse))))

; =======================
; Step logic
; =======================
; The mouse eats the tiger and no other eating relationships exist except those explicitly stated
; if if mouse eats tiger and if a eats tiger then  tiger visits mouse
(assert (animal_eats mouse tiger))
(assert (forall ((a Animal) (b Animal))
  (= (animal_eats a b)
     (and (= a mouse) (= b tiger)))))
(assert (=> (and (animal_eats mouse tiger)
                 (forall ((a Animal))
                   (=> (animal_eats a tiger)
                       (animal_visits tiger mouse))))
            (animal_visits tiger mouse)))
\end{lstlisting}
\end{tcolorbox}

\vspace{1mm}
\textbf{Evaluation:} \textcolor{red!70!black}{Fail due to Ungrounded.} \\
\textit{Reason:  ``no other
eating relationships exist except this explicitly stated'' cannot be grounded.}
\end{tcolorbox}

% ===== Step 3 =====
\begin{tcolorbox}[breakable, colback=white, colframe=black, title={Step 3}]
\scriptsize
\textbf{Step Context:} Since the tiger visits the mouse, and there is a rule stating that if something visits the mouse then the mouse is big, we can conclude that the mouse is big.

\vspace{2mm}
\begin{tcolorbox}[breakable, colback=black!2, colframe=black]
\begin{lstlisting}[style=solver]
; =======================
; Schema
; =======================
; (none new)

; =======================
; Added premises
; =======================
; (none new)

; =======================
; Step logic
; =======================
; if tiger visits mouse and if any animal visits mouse then mouse is big 
(assert (=> 
  (and (animal_visits tiger mouse)
       (forall ((a Animal))
         (=> (animal_visits a mouse)
             (animal_is_big mouse))))
  (animal_is_big mouse)))
\end{lstlisting}
\end{tcolorbox}

\vspace{1mm}
\textbf{Evaluation:} \textcolor{green!40!black}{Pass}
\end{tcolorbox}

% ===== Step 4 =====
\begin{tcolorbox}[breakable, colback=white, colframe=black, title={Step 4}]
\scriptsize
\textbf{Step Context:} Since the mouse is big, and there is a rule stating that if something is big then it visits the rabbit, we can conclude that the mouse visits the rabbit.

\vspace{2mm}
\begin{tcolorbox}[breakable, colback=black!2, colframe=black]
\begin{lstlisting}[style=solver]
; =======================
; Schema
; =======================
; constant representing the rabbit
(declare-const rabbit Animal)

; =======================
; Added premises
; =======================
; If something is big then it visits the rabbit (document)
(assert (forall ((a Animal))
  (=> (animal_is_big a)
      (animal_visits a rabbit))))

; =======================
; Step logic
; =======================
; if mouse big and if big then visits rabbit then mouse visits rabbit
(assert (=> 
  (and (animal_is_big mouse)
       (forall ((a Animal))
         (=> (animal_is_big a)
             (animal_visits a rabbit))))
  (animal_visits mouse rabbit)))
\end{lstlisting}
\end{tcolorbox}

\vspace{1mm}
\textbf{Evaluation:} \textcolor{green!40!black}{Pass}
\end{tcolorbox}

\begin{tcolorbox}[breakable, colback=white, colframe=black, title={Step 5}]
\scriptsize
\textbf{Step Context:} The statement 'The mouse visits the rabbit' is true based on the logical chain of the mouse being big and the rule about big things visiting the rabbit.

\vspace{2mm}
\begin{tcolorbox}[breakable, colback=black!2, colframe=black]
\begin{lstlisting}[style=solver]
; =======================
; Schema
; =======================
; (none new)

; =======================
; Added premises
; =======================
; (none new)

; =======================
; Step logic
; =======================
; if mouse is big and big animals visit rabbit then mouse visits rabbit
(assert (=> 
  (and (animal_is_big mouse)
       (forall ((a Animal))
         (=> (animal_is_big a)
             (animal_visits a rabbit))))
  (animal_visits mouse rabbit)))
\end{lstlisting}
\end{tcolorbox}

\vspace{1mm}
\textbf{Evaluation:} \textcolor{green!40!black}{Pass}
\end{tcolorbox}

\method{} Output after self-reflection:

\begin{tcolorbox}[breakable, colback=white, colframe=black, title={Decomposed CoT Steps}]
\scriptsize

\vspace{2mm}
1: From the reference information, we know three direct facts about the mouse: the mouse eats the tiger, the mouse is green, and there is a rule stating that if something visits the mouse then the mouse is big.

2: According to the reference information rule “If something eats the tiger then the tiger visits the mouse”, and since we know the mouse eats the tiger, we can conclude that the tiger visits the mouse.

3: Since we have established that the tiger visits the mouse, and we know from the reference information that “If something visits the mouse then the mouse is big”, we can conclude that the mouse is big.

4: The reference information states “If something is big then it visits the rabbit”. Since we have established that the mouse is big, applying this rule means that the mouse visits the rabbit.

5: Therefore, based on the logical chain starting from the mouse eating the tiger, leading to the tiger visiting the mouse, making the mouse big, and ending with the rule about big things visiting the rabbit, we can conclude that the statement “The mouse visits the rabbit” is true.

\end{tcolorbox}
% =======================
% Step 1
% =======================
\begin{tcolorbox}[breakable, colback=white, colframe=black, title={Step 1}]
\scriptsize
\textbf{Step Context:} From the reference information, we know three direct facts about the mouse: (1) The mouse eats the tiger, (2) The mouse is green, and (3) There is a rule stating that if something visits the mouse then the mouse is big.

\vspace{2mm}
\begin{tcolorbox}[breakable, colback=black!2, colframe=black]
\begin{lstlisting}[style=solver]
; =======================
; Schema
; =======================
; represents an animal entity
(declare-sort Animal)
; constant representing the mouse
(declare-const mouse Animal)
; constant representing the tiger
(declare-const tiger Animal)
; true if one animal eats another
(declare-fun animal_eats (Animal Animal) Bool)
; true if an animal is green
(declare-fun animal_is_green (Animal) Bool)
; true if an animal is big
(declare-fun animal_is_big (Animal) Bool)
; true if one animal visits another
(declare-fun animal_visits (Animal Animal) Bool)

; =======================
; Added premises
; =======================
; The mouse is green (document)
(assert 
  (animal_is_green mouse))

; The mouse and tiger are different animals (commonsense)
(assert 
  (not 
    (= mouse tiger)))
; The mouse eats the tiger (document)
; mouse eats tiger
(assert 
  (animal_eats mouse tiger))

; If something visits the mouse then the mouse is big (document)
; if a visits mouse then mouse is big
(assert 
  (forall 
    ((a Animal))
     (=> 
      (animal_visits a mouse) 
      (animal_is_big mouse))))

; =======================
; Step logic
; =======================
; mouse eats tiger
; mouse is green
; if animal visits mouse then mouse is big
(assert 
  (animal_eats mouse tiger))
(assert 
  (animal_is_green mouse))
(assert 
  (forall 
    ((a Animal))
     (=> 
      (animal_visits a mouse) 
      (animal_is_big mouse))))
\end{lstlisting}
\end{tcolorbox}

\vspace{1mm}
\textbf{Evaluation:} \textcolor{green!40!black}{Pass}
\end{tcolorbox}

% =======================
% Step 2
% =======================
\begin{tcolorbox}[breakable, colback=white, colframe=black, title={Step 2}]
\scriptsize
\textbf{Step Context:} According to the reference information rule 'If something eats the tiger then the tiger visits the mouse', and since we know the mouse eats the tiger, we can conclude that the tiger visits the mouse.

\vspace{2mm}
\begin{tcolorbox}[breakable, colback=black!2, colframe=black]
\begin{lstlisting}[style=solver]
; =======================
; Schema
; =======================
; (none new)

; =======================
; Added premises
; =======================
; If something eats the tiger then the tiger visits the mouse (document)
; if a eats tiger then tiger visits mouse
(assert 
  (forall 
    ((a Animal))
     (=> 
      (animal_eats a tiger) 
      (animal_visits tiger mouse))))

; =======================
; Step logic
; =======================
; if a eats tiger then tiger visits mouse
; if mouse eats tiger then tiger visits mouse
(assert 
  (forall 
    ((a Animal))
     (=> 
      (animal_eats a tiger) 
      (animal_visits tiger mouse))))
(assert 
  (=> 
    (animal_eats mouse tiger) 
    (animal_visits tiger mouse)))
\end{lstlisting}
\end{tcolorbox}

\vspace{1mm}
\textbf{Evaluation:} \textcolor{green!40!black}{Pass}
\end{tcolorbox}

% =======================
% Step 3
% =======================
\begin{tcolorbox}[breakable, colback=white, colframe=black, title={Step 3}]
\scriptsize
\textbf{Step Context:} Since we have established that the tiger visits the mouse, and we know from the reference information that 'If something visits the mouse then the mouse is big', we can conclude that the mouse is big.

\vspace{2mm}
\begin{tcolorbox}[breakable, colback=black!2, colframe=black]
\begin{lstlisting}[style=solver]
; =======================
; Schema
; =======================
; (none new)

; =======================
; Added premises
; =======================
; (none new)

; =======================
; Step logic
; =======================
; if tiger visits mouse then mouse is big
(assert 
  (=> 
    (animal_visits tiger mouse) 
    (animal_is_big mouse)))
\end{lstlisting}
\end{tcolorbox}

\vspace{1mm}
\textbf{Evaluation:} \textcolor{green!40!black}{Pass}
\end{tcolorbox}

% =======================
% Step 4
% =======================
\begin{tcolorbox}[breakable, colback=white, colframe=black, title={Step 4}]
\scriptsize
\textbf{Step Context:} The reference information states 'If something is big then it visits the rabbit'. Since we have established that the mouse is big, applying this rule means that the mouse visits the rabbit.

\vspace{2mm}
\begin{tcolorbox}[breakable, colback=black!2, colframe=black]
\begin{lstlisting}[style=solver]
; =======================
; Schema
; =======================
; constant representing the rabbit
(declare-const rabbit Animal)

; =======================
; Added premises
; =======================
;If something is big then it visits the rabbit (document)
; if a is big then a visits rabbit
(assert 
  (forall 
    ((a Animal))
     (=> 
      (animal_is_big a) 
      (animal_visits a rabbit))))

; =======================
; Step logic
; =======================
; if a is big then a visits rabbit
; if mouse is big then mouse visits rabbit
(assert 
  (forall 
    ((a Animal))
     (=> 
      (animal_is_big a) 
      (animal_visits a rabbit))))
(assert 
  (=> 
    (animal_is_big mouse) 
    (animal_visits mouse rabbit)))
\end{lstlisting}
\end{tcolorbox}

\vspace{1mm}
\textbf{Evaluation:} \textcolor{green!40!black}{Pass}
\end{tcolorbox}

% =======================
% Step 5
% =======================
\begin{tcolorbox}[breakable, colback=white, colframe=black, title={Step 5}]
\scriptsize
\textbf{Step Context:} Therefore, based on the logical chain starting from the mouse eating the tiger, leading to the tiger visiting the mouse, making the mouse big, and ending with the rule about big things visiting the rabbit, we can conclude that the statement 'The mouse visits the rabbit' is true.

\vspace{2mm}
\begin{tcolorbox}[enhanced jigsaw,               % required for reliable breaking with complex content
  breakable,                     % allow the box to split across pages
  verbatim,                      % <<— allows verbatim content (lstlisting) inside
  colback=black!2,
  colframe=black,]
\begin{lstlisting}[style=solver]
; =======================
; Schema
; =======================
; (none new)

; =======================
; Added premises
; =======================
; (none new)

; =======================
; Step logic
; =======================
; mouse eats tiger => tiger visits mouse
; tiger visits mouse => mouse is big
; mouse is big => mouse visits rabbit
; mouse visits rabbit
(assert 
  (=> 
    (animal_eats mouse tiger) 
    (animal_visits tiger mouse)))
(assert 
  (=> 
    (animal_visits tiger mouse) 
    (animal_is_big mouse)))
(assert 
  (=> 
    (animal_is_big mouse) 
    (animal_visits mouse rabbit)))
(assert 
  (animal_visits mouse rabbit))
\end{lstlisting}
\end{tcolorbox}

\vspace{1mm}
\textbf{Evaluation:} \textcolor{green!40!black}{Pass}
\end{tcolorbox}

\subsubsection{Contradiction example from BioASQ}
\label{appendix:exmaple_bioasq}

\begin{tcolorbox}[breakable, colback=white, colframe=black, title={Contradiction example from BioASQ}]
\scriptsize
\textbf{Question:} Can Connexin hemi channels be used for drug delivery? \\[1mm]
\textbf{Gold Answer:} Yes

\vspace{2mm}
\textbf{Context:} \\
The permeability of Cx43 channels to small molecules and macromolecules makes them highly attractive targets for delivering drugs directly into the cytoplasm. Cancer cells overexpressing Cx43 may be more permeable and sensitive to chemotherapeutics. In this context, certain channels lead to transitory plasma membrane permeability changes, such as pannexin, connexin hemichannels that are channels in membranes that pass molecules, TRPV1-4 and P2×7, which allow for the non-selective passage of molecules up to 1,000 Da.

Gap junction channels, composed of connexin proteins, provide a mechanism for direct transfer of small molecules across membranes, and recent evidence suggests that the transfer of larger, polymer-like molecules such as microRNAs may be possible. 

Collectively, these results reveal that polymeric macromolecules can be delivered to cells via gap junctions, suggesting that the gap junction route can be used for the delivery of macro polymeric therapeutic molecules, which provides evidence for drug delivery potential.

\vspace{2mm}
\textbf{Original CoT:} \\
Let me break this down step by step.\\[1mm]

1. First, let's understand what connexin hemichannels are:

* They are channels in cell membranes that can allow molecules to pass through.  

* They are part of a gap junction.  

2. Key characteristic of connexin hemichannels:  

* They allow "non-selective passage of molecules up to 1,000 Da".  

3. Evidence for drug delivery potential:  

* Gap junction route can transport various types of macro polymeric therapeutic molecules.  

\vspace{1mm}
\textbf{Conclusion:} Yes, connexin hemichannels can be used for drug delivery. The evidence strongly supports this, showing they can transport various types of therapeutic molecules, making them valuable targets for drug delivery applications.
\end{tcolorbox}

\begin{tcolorbox}[breakable, colback=white, colframe=black, title={Decomposed CoT Steps}]
\scriptsize
1. Connexin hemichannels are channels in cell membranes that can allow molecules to pass through and are part of gap junction channels. \\[1mm]
2. Connexin hemichannels allow for the non-selective passage of molecules up to 1,000 Da. \\[1mm]
3. The gap junction route can transport macro polymer therapeutic molecules, which is evidence for drug delivery potential. \\[1mm]
4. Therefore, Connexin hemi channels are permeable to therapeutic macromolecules, makes them highly attractive targets for delivering drugs.
\end{tcolorbox}

% =======================
% Step 1
% =======================
\begin{tcolorbox}[breakable, colback=white, colframe=black, title={Step 1}]
\scriptsize
\textbf{Step Context:} Connexin hemichannels are channels in cell membranes that can allow molecules to pass through and are part of gap junction channels.

\vspace{2mm}
\begin{tcolorbox}[breakable, colback=black!2, colframe=black]
\begin{lstlisting}[style=solver]
; =======================
; Schema
; =======================
; represents a molecule
(declare-sort Molecule)
; represents a channel
(declare-sort Channel)
; whether a channel is a connexin hemichannel
(declare-fun IsConnexinHemichannel (Channel) Bool)
; whether a channel is a gap junction channel
(declare-fun IsGapJunctionChannel (Channel) Bool)
; whether a channel is located in a cell membrane
(declare-fun IsInCellMembrane (Channel) Bool)
; whether an entity is considered a channel
(declare-fun IsChannel (Channel) Bool)
; whether a molecule can enter through a channel
(declare-fun CanEnterViaChannel (Molecule Channel) Bool)
; whether one channel is part of another channel
(declare-fun PartOf (Channel Channel) Bool)
; =======================
; Added premises
; =======================
; Connexin hemichannels are channels in membranes that pass molecules (document)
(assert (forall ((c Channel))
  (=> (IsConnexinHemichannel c)
      (and (IsChannel c)
           (IsInCellMembrane c)
           (exists ((m Molecule)) (CanEnterViaChannel m c)))))))

; Connexin hemichannels are part of gap junction channels (document)
(assert (forall ((h Channel))
  (=> (IsConnexinHemichannel h)
      (exists ((g Channel))
        (and (IsGapJunctionChannel g)
             (PartOf h g))))))

; =======================
; Step logic
; =======================
; Connexin hemichannels are channels in membranes that pass molecules
; Connexin hemichannels are part of gap junction channels
(assert (forall ((c Channel))
  (=> (IsConnexinHemichannel c)
      (and (IsChannel c)
           (IsInCellMembrane c)
           (exists ((m Molecule)) (CanEnterViaChannel m c)))))))
(assert (forall ((h Channel))
  (=> (IsConnexinHemichannel h)
      (exists ((g Channel))
        (and (IsGapJunctionChannel g)
             (PartOf h g))))))
\end{lstlisting}
\end{tcolorbox}

\vspace{1mm}
\textbf{Evaluation:} \textcolor{green!40!black}{Pass}
\end{tcolorbox}

% =======================
% Step 2
% =======================
\begin{tcolorbox}[breakable, colback=white, colframe=black, title={Step 2}]
\scriptsize
\textbf{Step Context:} Connexin hemichannels allow for the non-selective passage of molecules up to 1,000 Da.

\vspace{2mm}
\begin{tcolorbox}[breakable, colback=black!2, colframe=black]
\begin{lstlisting}[style=solver]
; =======================
; Schema
; =======================
; returns the molecular weight of a molecule
(declare-fun MolecularWeight (Molecule) Real)

; =======================
; Added premises
; =======================
; Connexin hemichannels allow for the non-selective passage of molecules up to 1,000 Da (document)
(assert (forall ((m Molecule) (c Channel))
  (=> (and (IsConnexinHemichannel c)
           (CanEnterViaChannel m c))
      (<= (MolecularWeight m) 1000))))

; =======================
; Step logic
; =======================
; If a molecule can enter via connexin channel, its weight must be <= 1000
(assert (forall ((m Molecule) (c Channel))
  (=> (and (IsConnexinHemichannel c)
           (CanEnterViaChannel m c))
      (<= (MolecularWeight m) 1000))))
\end{lstlisting}
\end{tcolorbox}

\vspace{1mm}
\textbf{Evaluation:} \textcolor{green!40!black}{Pass}
\end{tcolorbox}

% =======================
% Step 3
% =======================
\begin{tcolorbox}[breakable, colback=white, colframe=black, title={Step 3}]
\scriptsize
\textbf{Step Context:} Gap junction route can transport macro polymer therapeutic molecules, which is evidence for drug delivery potential.

\vspace{2mm}
\begin{tcolorbox}[breakable, colback=black!2, colframe=black]
\begin{lstlisting}[style=solver]
; =======================
; Schema
; =======================
; whether a molecule is a macro polymer therapeutic molecule
(declare-fun IsMacroPolymerTherapeuticmolecule (Molecule) Bool)
; whether a channel has potential for drug delivery
(declare-fun HasDrugDeliveryPotential (Channel) Bool)
; =======================
; Added premises
; =======================
; The ability to deliver macro polymeric therapeutic molecules via the gap junction route,
; which provides evidence for drug delivery potential (document)
(assert
  (forall ((c Channel) (m Molecule))
    (=> (and (IsGapJunctionChannel c)
             (IsMacroPolymerTherapeuticmolecule m))
        (and (CanEnterViaChannel m c)
             (=> (CanEnterViaChannel m c)
                 (HasDrugDeliveryPotential c))))))

; Any marco polymeric therapeutic molecule has a molecular weight > 1,000 Da (commonsense)
(assert
  (forall ((m Molecule))
    (=> (IsMacroPolymerTherapeuticmolecule m)
        (> (MolecularWeight m) 1000))))

; =======================
; Step logic
; =======================
; If a gap junction channel transports large polymeric therapeutic macromolecules,
; then it has drug-delivery potential
(assert
  (forall ((c Channel) (m Molecule))
    (=> (and (IsGapJunctionChannel c)
             (IsMacroPolymerTherapeuticmolecule m))
        (and (CanEnterViaChannel m c)
             (=> (CanEnterViaChannel m c)
                 (HasDrugDeliveryPotential c))))))
\end{lstlisting}
\end{tcolorbox}

\vspace{1mm}
\textbf{Evaluation:} \textcolor{green!40!black}{Pass}
\end{tcolorbox}

% =======================
% Step 4 (Conclusion)
% =======================
\begin{tcolorbox}[breakable, colback=white, colframe=black, title={Step 4}]
\scriptsize
\textbf{Step Context:} Therefore, Connexin hemichannels are permeable to therapeutic macromolecules, making them valuable targets for delivering drugs.

\vspace{2mm}
\begin{tcolorbox}[breakable, colback=black!2, colframe=black]
\begin{lstlisting}[style=solver]
; =======================
; Schema
; =======================
; (none new)

; =======================
; Added premises
; =======================
; (none new)

; =======================
; Step logic
; =======================
; Connexin hemichannels are permeable to therapeutic macromolecules
(assert
  (exists ((c Channel) (m Molecule))
    (and (IsConnexinHemichannel c)
         (IsMacroPolymerTherapeuticmolecule m)
         (CanEnterViaChannel m c))))

; Connexin hemichannels are permeable to therapeutic macromolecules,
; -> they are highly attractive drug-delivery targets
(assert
  (forall ((c Channel) (m Molecule))
    (=> (and (IsConnexinHemichannel c)
             (IsMacroPolymerTherapeuticmolecule m)
             (CanEnterViaChannel m c))
        (HasDrugDeliveryPotential c))))
\end{lstlisting}
\end{tcolorbox}

\vspace{1mm}
\textbf{Evaluation:} \textcolor{red!70!black}{Fail due to contradiction.} \\
\textit{Reason:} The axioms show macro-polymer molecules are $> 1000$ Da, but connexin hemichannels only permit molecules $\leq 1000$ Da. Thus, the same molecule would need to be both $\leq 1000$ and $> 1000$, which is impossible.
\end{tcolorbox}

\begin{tcolorbox}[breakable, colback=white, colframe=black, title={Decomposed CoT Steps}]
\scriptsize
1. Connexin hemichannels are channels in cell membranes that can allow molecules to pass through and are part of gap junction channels. \\[1mm]
2. Connexin hemichannels allow for the non-selective passage of molecules up to 1,000 Da. \\[1mm]
3. Cx43 channels are permeable to both small molecules and macromolecules, makes them highly attractive targets for delivering drugs. \\[1mm]
4. Since Cx43 is a connexin hemi channel and supports drug delivery, we infer that connexin hemi channels can be used for drug delivery.
\end{tcolorbox}

% =======================
% Step 1
% =======================
\begin{tcolorbox}[breakable, colback=white, colframe=black, title={Step 1}]
\scriptsize
\textbf{Step Context:} connexin hemi channel are channels in cell membranes that can allow molecules to pass through and are part of a gap junction channels.

\vspace{2mm}
\begin{tcolorbox}[breakable, colback=black!2, colframe=black]
\begin{lstlisting}[style=solver]
; =======================
; Schema
; =======================
; represents a molecule
(declare-sort Molecule)
; represents a channel
(declare-sort Channel)
; whether a channel is a connexin hemichannel
(declare-fun IsConnexinHemichannel (Channel) Bool)
; whether a channel is a gap junction channel
(declare-fun IsGapJunctionChannel (Channel) Bool)
; whether a channel is located in a cell membrane
(declare-fun IsInCellMembrane (Channel) Bool)
; whether an entity is considered a channel
(declare-fun IsChannel (Channel) Bool)
; whether a molecule can enter through a channel
(declare-fun CanEnterViaChannel (Molecule Channel) Bool)
; whether one channel is part of another channel
(declare-fun PartOf (Channel Channel) Bool)
; =======================
; Added premises
; =======================
; connexin hemmichannels are channels in membranes that pass molecules( documents)
(assert (forall ((c Channel))
  (=> (IsConnexinHemichannel c)
      (and (IsChannel c)
           (IsInCellMembrane c)
           (exists ((m Molecule)) (CanEnterViaChannel m c))))))

; Connexin hemichannels are part of gap junction channels ( documents)
(assert (forall ((h Channel))
  (=> (IsConnexinHemichannel h)
      (exists ((g Channel))
        (and (IsGapJunctionChannel g)
             (PartOf h g))))))

; =======================
; Step logic
; =======================
; Connexin hemichannels are channels in membranes that pass molecules
(assert (forall ((c Channel))
  (=> (IsConnexinHemichannel c)
      (and (IsChannel c)
           (IsInCellMembrane c)
           (exists ((m Molecule)) (CanEnterViaChannel m c))))))

; Connexin hemichannels are part of gap junction channels
(assert (forall ((h Channel))
  (=> (IsConnexinHemichannel h)
      (exists ((g Channel))
        (and (IsGapJunctionChannel g)
             (PartOf h g))))))
\end{lstlisting}
\end{tcolorbox}

\vspace{1mm}
\textbf{Evaluation:} pass
\end{tcolorbox}

% =======================
% Step 2
% =======================
\begin{tcolorbox}[breakable, colback=white, colframe=black, title={Step 2}]
\scriptsize
\textbf{Step Context:} Connexin hemi channels allow for the non-selective passage of molecules up to 1,000 Da.

\vspace{2mm}
\begin{tcolorbox}[breakable, colback=black!2, colframe=black]
\begin{lstlisting}[style=solver]
; =======================
; Schema
; =======================
; returns the molecular weight of a molecule
(declare-fun MolecularWeight (Molecule) Real)
; =======================
; Added premises
; =======================
; Connexin hemmichannels allow for the non-selective passage of molecules up to 1,000 Da. 
(assert (forall ((m Molecule) (c Channel))
  (=> (and 
        (IsConnexinHemichannel c)
        (CanEnterViaChannel m c))
      (<= (MolecularWeight m) 1000))))

; =======================
; Step logic
; =======================
; If a drug can enter via connexin channel, its weight must be <= 1000
(assert (forall ((m Molecule) (c Channel))
  (=> (and 
        (IsConnexinHemichannel c)
        (CanEnterViaChannel m c))
      (<= (MolecularWeight m) 1000))))
\end{lstlisting}
\end{tcolorbox}

\vspace{1mm}
\textbf{Evaluation:} \textcolor{green!40!black}{Pass}
\end{tcolorbox}
% =======================
% Step 3
% =======================
\begin{tcolorbox}[breakable, colback=white, colframe=black, title={Step 3}]
\scriptsize
\textbf{Step Context:} Cx43 channels are permeable to both small molecules and macromolecules, makes them highly attractive targets for delivering drugs.

\vspace{2mm}
\begin{tcolorbox}[breakable, colback=black!2, colframe=black]
\begin{lstlisting}[style=solver]
; =======================
; Schema
; =======================
; whether a channel is specifically connexin 43 (Cx43)
(declare-fun IsCx43 (Channel) Bool)
; whether a molecule is a small molecule
(declare-fun IsSmallMolecule (Molecule) Bool)
; whether a molecule is a macromolecule
(declare-fun IsMacromolecule (Molecule) Bool)
; whether a channel is permeable to a given molecule
(declare-fun PermeableTo (Channel Molecule) Bool)
; whether a channel can deliver drugs (channel-level, independent of molecule)
(declare-fun CanDeliverDrug (Channel) Bool)
; =======================
; Added premises
; =======================
; The permeability of Cx43 channels to small molecules and macromolecules, makes them highly attractive targets for delivering drugs directly into the cytoplasm. (document)
(assert
  (forall ((c Channel) (m Molecule))
    (=> (and (IsCx43 c)
             (or (IsSmallMolecule m)
                 (IsMacromolecule m)))
        (and (CanEnterViaChannel m c)
             (=> (CanEnterViaChannel m c)
                 (CanDeliverDrug c))))))

; Any large macromolecule has a molecular weight > 1,000 Da. (commonsense)
(assert
  (forall ((m Molecule))
    (=> (IsMacromolecule m)
        (> (MolecularWeight m) 1000))))

; Any small macromolecule has a molecular weight <= 1,000 Da. (commonsense)
(assert
  (forall ((m Molecule))
    (=> (IsSmallMolecule m)
        (<= (MolecularWeight m) 1000))))

; =======================
; Step logic
; =======================
; Combined: if c is Cx43 and m is small or a macromolecule, then
; (1) m permeates via c; and (2) given that permeability, c can deliver drugs.
(assert
  (forall ((c Channel) (m Molecule))
    (=> (and (IsCx43 c)
             (or (IsSmallMolecule m)
                 (IsMacromolecule m)))
        (and (CanEnterViaChannel m c)
             (=> (CanEnterViaChannel m c)
                 (CanDeliverDrug c))))))
\end{lstlisting}
\end{tcolorbox}

\vspace{1mm}
\textbf{Evaluation:} \textcolor{green!40!black}{Pass}
\end{tcolorbox}

% =======================
% Step 4
% =======================
\begin{tcolorbox}[breakable, colback=white, colframe=black, title={Step 4}]
\scriptsize
\textbf{Step Context:} Since Cx43 is a connexin hemi channel and supports drug delivery, we infer that connexin hemi channels can be used for drug delivery.

\vspace{2mm}
\begin{tcolorbox}[breakable, colback=black!2, colframe=black]
\begin{lstlisting}[style=solver]
; =======================
; Schema
; =======================
; (none new)

; =======================
; Added premises
; =======================
; Cx43 channels are a type of connexin channel (Commonsense)
; Cx43 is a type of connexin hemichannel
(assert (forall ((c Channel))
  (=> (IsCx43 c)
      (IsConnexinHemichannel c))))

; =======================
; Step logic
; =======================
; Cx43 channels are a type of connexin channel
; Cx43 is a type of connexin hemichannel
(assert (forall ((c Channel))
  (=> (IsCx43 c)
      (IsConnexinHemichannel c))))

; If a specific Cx43 can deliver drugs, infer that connexin hemichannels can.
; Every drug that a Cx43 channel can deliver, can also be delivered by some Connexin hemichannel.

(assert
  (exists ((c Channel))
    (and (IsCx43 c)
         (CanDeliverDrug c))))

(assert
  (forall ((c Channel))
    (=> (and (IsCx43 c)
             (CanDeliverDrug c))
        (exists ((h Channel))
          (and (IsConnexinHemichannel h)
               (CanDeliverDrug h))))))
\end{lstlisting}
\end{tcolorbox}

\vspace{1mm}
\textbf{Evaluation:} \textcolor{green!40!black}{Pass}
\end{tcolorbox}

\subsubsection{Untranslatable example from SARA}
\label{appendix:example_sara}

\begin{tcolorbox}[breakable, colback=white, colframe=black, title={Untranslatable example from SARA}]
\scriptsize
\textbf{Reference information:} Section 3306(c)(1) applies to agricultural labor.  

\vspace{2mm}
\textbf{Question:} Determine whether the following statements are entailed under the statute.  
Description: Alice has paid \$3200 to Bob for agricultural labor done from Feb 1st, 2017 to Sep 2nd, 2017. Alice is an American employer.  
Statement: Section 3306(c)(1) applies to Alice employing Bob for the year 2017.  
Reply with either: Entailment, Contradiction.\\[1mm]

\textbf{Answer:} Entailment  

\vspace{2mm}
\textbf{Original CoT:} \\
Let me solve this step by step.\\[1mm]

First, let's identify the key elements in the statute:  

* The statute in question is Section 3306(c)(1).  

* Its primary subject is "agricultural labor".  

Let's analyze the given description of the situation:  

* Employee: Bob  

* Employer: Alice (an American employer)  

* Type of work: The work is explicitly identified as 
agricultural labor.  

Let's compare the situation to the statute's basic scope:  

* The statute, Section 3306(c)(1), specifically and directly addresses agricultural labor.  

* The work performed by Bob is agricultural labor.  

* Therefore, the situation falls squarely within the fundamental subject matter of this statutory section.  

Important consideration:  

* The fact that the work was performed for an American employer raises potential jurisdictional or geographical questions.  

* However, the provided statute fragment does not contain any information about geographical limitations.  

Based on the basic provision:  

* Since the work is agricultural labor, and Section 3306(c)(1) explicitly covers agricultural labor, the situation described falls under the basic scope of this section.  
\end{tcolorbox}

\method{} Output before self-reflection:
\begin{tcolorbox}[breakable, colback=white, colframe=black, title={Decomposed CoT Steps}]
\scriptsize
1. Section 3306(c)(1) of the statute addresses agricultural labor

2. The work performed by Bob for Alice is specifically agricultural labor

3. The basic scope of Section 3306(c)(1) covers all agricultural labor

4. While Alice is an American employer and there might be jurisdictional implications, the statute fragment does not provide information about geographical limitations

5. Bob's employment falls under the basic scope of Section 3306(c)(1)
\end{tcolorbox}
\begin{tcolorbox}[breakable, colback=white, colframe=black, title={Step 1}]
\scriptsize
\textbf{Step Context:} Section 3306(c)(1) of the statute addresses agricultural labor

\vspace{2mm}
\begin{tcolorbox}[breakable, colback=black!2, colframe=black]
\begin{lstlisting}[style=solver]
; =======================
; Schema:
; =======================
; represents a section of statute
(declare-sort StatuteSection)

; specific section 3306(c)(1)
(declare-const section_3306c1 StatuteSection)
; whether section deals with agricultural labor
(declare-fun deals_with_agricultural_labor (StatuteSection) Bool)

; =======================
; Add premises:
; =======================
; Section 3306(c)(1) applies to agricultural labor (document)
(assert 
  (deals_with_agricultural_labor section_3306c1))

; =======================
; Step Logic: 
; =======================
; sec 3306c1 abt agri labor
(assert 
  (deals_with_agricultural_labor section_3306c1))
\end{lstlisting}
\end{tcolorbox}

\vspace{1mm}
\textbf{Evaluation:} \textcolor{green!40!black}{Pass}
\end{tcolorbox}
\begin{tcolorbox}[breakable, colback=white, colframe=black, title={Step 2}]
\scriptsize
\textbf{Step Context:} The work performed by Bob for Alice is specifically agricultural labor

\vspace{2mm}
\begin{tcolorbox}[breakable, colback=black!2, colframe=black]
\begin{lstlisting}[style=solver]
; =======================
; Add schema:
; =======================
; represents a person
(declare-sort Person)
; represents Alice
(declare-const alice Person)
; represents Bob
(declare-const bob Person)
; whether person performs agricultural labor
(declare-fun performs_agricultural_labor (Person) Bool)
; whether person employs another person
(declare-fun employs (Person Person) Bool)

; =======================
; Add premises:
; =======================
; Alice employed Bob to perform agricultural labor from February 1st to September 2nd, 2017, as evidenced by Alice's payment of $3200 for this work (document)
; Dates encoded as YYYYMMDD
; paid employer worker laborType amount startDate endDate
(declare-fun paid (Person Person LaborType Int Int Int) Bool)

(assert 
  (and 
    (employs alice bob)
    (performs_agricultural_labor bob) 
    (paid alice bob agricultural_labor 3200 20170201 20170902)))

; =======================
; Step logic:
; =======================
; Bob performs agri labor for Alice
(assert 
  (and 
    (performs_agricultural_labor bob) 
    (employs alice bob)))
\end{lstlisting}
\end{tcolorbox}

\vspace{1mm}
\textbf{Evaluation:} \textcolor{green!40!black}{Pass}
\end{tcolorbox}

% =======================
% Step 3
% =======================
\begin{tcolorbox}[breakable, colback=white, colframe=black, title={Step 3}]
\scriptsize
\textbf{Step Context:} The basic scope of Section 3306(c)(1) covers all agricultural labor

\vspace{2mm}
\begin{tcolorbox}[breakable, colback=black!2, colframe=black]
\begin{lstlisting}[style=solver]
; =======================
; Add schema:
; =======================
; predicate: whether a person's work is covered by 3306(c)(1)
(declare-fun covered_by_section_3306c1 (Person) Bool)

; =======================
; Add premises:
; =======================
; A person performs labor, therefore anyone perform agricultural labor is covered by Section 3306(c)(1) (commonsense)
(assert
  (forall ((p Person))
    (=> (performs_agricultural_labor p)
        (covered_by_section_3306c1 p))))

; =======================
; Step logic:
; =======================
; sec 3306c1 covers agri labor
; if person does agri labor then covered by sec 3306c1
(assert 
  (deals_with_agricultural_labor section_3306c1))
(assert
  (forall ((p Person))
    (=> (performs_agricultural_labor p)
        (covered_by_section_3306c1 p))))
\end{lstlisting}
\end{tcolorbox}

\vspace{1mm}
\textbf{Evaluation:} \textcolor{green!40!black}{Pass}
\end{tcolorbox}

% =======================
% Step 4
% =======================
\begin{tcolorbox}[breakable, colback=white, colframe=black, title={Step 4}]
\scriptsize
\textbf{Step Context:} While Alice is an American employer and there might be jurisdictional implications, the statute fragment does not provide information about geographical limitations

\vspace{2mm}
\begin{tcolorbox}[breakable, colback=black!2, colframe=black]
\begin{lstlisting}[style=solver]
; =======================
; Schema:
; (none new)
; =======================

; =======================
; Add premises:
; (none new)
; =======================

; =======================
; Step logic:
; (none new)
; =======================
\end{lstlisting}
\end{tcolorbox}

\vspace{1mm}
\textbf{Evaluation:} \textcolor{red!70!black}{Fail due to Untranslatable} \\
\textit{Reason: The phrase ``there might be jurisdictional implication'' cannot be rendered under SMI-LIB rules because it expresses a possibility rather than a definitive statement.”}
\end{tcolorbox}

% =======================
% Step 5
% =======================
\begin{tcolorbox}[breakable, colback=white, colframe=black, title={Step 5}]
\scriptsize
\textbf{Step Context:} Bob's employment falls under the basic scope of Section 3306(c)(1)

\vspace{2mm}
\begin{tcolorbox}[breakable, colback=black!2, colframe=black]
\begin{lstlisting}[style=solver]
; =======================
; Schema:
; (none new)
; =======================

; =======================
; Add premises:
; (none new)
; =======================

; =======================
; Step logic:
; =======================
(assert  (covered_by_section_3306c1 bob))
\end{lstlisting}
\end{tcolorbox}

\vspace{1mm}
\textbf{Evaluation:} \textcolor{green!40!black}{Pass}
\end{tcolorbox}

\method{} Output after self-reflection:

\begin{tcolorbox}[breakable, colback=white, colframe=black, title={Decomposed CoT steps}]
\scriptsize
1. Section 3306(c)(1) explicitly applied to agricultural labor as shown by the statute text 'agricultural labor'.

2. Bob performed agricultural labor.

3. Since Bob's work qualifies as agricultural labor, and Section 3306(c)(1) covers agricultural labor, Bob's employment falls under the basic scope of Section 3306(c)(1).

\end{tcolorbox}

\begin{tcolorbox}[breakable, colback=white, colframe=black, title={Step 1}]
\scriptsize
\textbf{Step Context:} Section 3306(c)(1) explicitly addresses agricultural labor, which means that any person who performs agricultural labor is covered by Section 3306(c)(1).

\vspace{2mm}
\begin{tcolorbox}[breakable, colback=black!2, colframe=black]
\begin{lstlisting}[style=solver]
; =======================
; Schema
; =======================
; represents a statute section
(declare-sort StatuteSection)
; represents a type of labor
(declare-sort LaborType)
; represents a person
(declare-sort Person)
; specific statute section 3306(c)(1)
(declare-const section_3306c1 StatuteSection)
; specific labor type: agricultural labor
(declare-const labor_agricultural LaborType)
; whether a section addresses a labor type
(declare-fun section_addresses_labor (StatuteSection LaborType) Bool)
; whether a person performed a labor type
(declare-fun performed_labor (Person LaborType) Bool)
; whether a person's employment is covered by a statute section
(declare-fun employment_covered (Person StatuteSection) Bool)

; =======================
; Added premises
; =======================
; Section 3306(c)(1) explicitly applies to agricultural labor as shown by the statute text 'agricultural labor' (document)
(assert (section_addresses_labor section_3306c1 labor_agricultural))

; A person performs labor, therefore anyone performing agricultural labor is covered by Section 3306(c)(1) (commonsense)
(assert
  (forall ((p Person))
    (=> (performed_labor p labor_agricultural)
        (employment_covered p section_3306c1))))

; =======================
; Step logic
; =======================
; Section 3306(c)(1) explicitly addresses agricultural labor
; -> anyone who performed agricultural labor is covered
(assert (section_addresses_labor section_3306c1 labor_agricultural))
(assert
  (forall ((p Person))
    (=> (performed_labor p labor_agricultural)
        (employment_covered p section_3306c1))))
\end{lstlisting}
\end{tcolorbox}

\vspace{1mm}
\textbf{Evaluation:} \textcolor{green!40!black}{Pass}
\end{tcolorbox}

% =======================
% Step 2
% =======================
\begin{tcolorbox}[breakable, colback=white, colframe=black, title={Step 2}]
\scriptsize
\textbf{Step Context:} Bob performed agricultural labor.

\vspace{2mm}
\begin{tcolorbox}[breakable, colback=black!2, colframe=black]
\begin{lstlisting}[style=solver]
; =======================
; Schema
; =======================
; Bob performed agricultural labor.
(declare-const Bob   Person)

; =======================
; Added premises
; =======================
; Alice has paid $3200 to Bob for agricultural labor done from Feb 1st, 2017 to Sep 2nd, 2017. (document)
; Dates encoded as YYYYMMDD
; paid employer worker laborType amount startDate endDate
(declare-fun paid (Person Person LaborType Int Int Int) Bool)
(declare-const Alice Person)
(assert (performed_labor Bob labor_agricultural))
(assert (paid Alice Bob labor_agricultural 3200 20170201 20170902))

; =======================
; Step logic
; =======================
(assert (performed_labor Bob labor_agricultural))
\end{lstlisting}
\end{tcolorbox}

\vspace{1mm}
\textbf{Evaluation:} \textcolor{green!40!black}{Pass}
\end{tcolorbox}

% =======================
% Step 3
% =======================
\begin{tcolorbox}[breakable, colback=white, colframe=black, title={Step 3}]
\scriptsize
\textbf{Step Context:} Since Bob's work qualifies as agricultural labor, and Section 3306(c)(1) covers agricultural labor, Bob's employment falls under the basic scope of Section 3306(c)(1).

\vspace{2mm}
\begin{tcolorbox}[breakable, colback=black!2, colframe=black]
\begin{lstlisting}[style=solver]
; =======================
; Schema
; (none new)
; =======================

; =======================
; Added premises
; (none new)
; =======================

; =======================
; Step logic
; =======================
; derive Bob's coverage
(assert (=> 
  (and
    (performed_labor Bob labor_agricultural)
    (forall ((p Person))
      (=> (performed_labor p labor_agricultural)
          (employment_covered p section_3306c1))))
  (employment_covered Bob section_3306c1)))

\end{lstlisting}
\end{tcolorbox}

\vspace{1mm}
\textbf{Evaluation:} \textcolor{green!40!black}{Pass}
\end{tcolorbox}

%% file: iclr2026_conference.bbl
\begin{thebibliography}{37}
\providecommand{\natexlab}[1]{#1}
\providecommand{\url}[1]{\texttt{#1}}
\expandafter\ifx\csname urlstyle\endcsname\relax
  \providecommand{\doi}[1]{doi: #1}\else
  \providecommand{\doi}{doi: \begingroup \urlstyle{rm}\Url}\fi

\bibitem[{Anthropic}(2024)]{anthropic2024claude3_5_sonnet}
{Anthropic}.
\newblock Introducing claude 3.5 sonnet.
\newblock Anthropic press release, Claude.ai, June~20 2024.
\newblock Accessed 2025-08-09.

\bibitem[Barrett et~al.(2016)Barrett, Fontaine, and Tinelli]{BarFT-SMTLIB}
Clark Barrett, Pascal Fontaine, and Cesare Tinelli.
\newblock {The Satisfiability Modulo Theories Library (SMT-LIB)}.
\newblock {\tt www.SMT-LIB.org}, 2016.

\bibitem[Bender \& Koller(2020)Bender and Koller]{bender-koller-2020-climbing}
Emily~M. Bender and Alexander Koller.
\newblock Climbing towards {NLU}: {On} meaning, form, and understanding in the
  age of data.
\newblock In Dan Jurafsky, Joyce Chai, Natalie Schluter, and Joel Tetreault
  (eds.), \emph{Proceedings of the 58th Annual Meeting of the Association for
  Computational Linguistics}, pp.\  5185--5198, Online, July 2020. Association
  for Computational Linguistics.
\newblock \doi{10.18653/v1/2020.acl-main.463}.
\newblock URL \url{https://aclanthology.org/2020.acl-main.463/}.

\bibitem[Bostrom et~al.(2021)Bostrom, Zhao, Chaudhuri, and
  Durrett]{bostrom-etal-2021-flexible}
Kaj Bostrom, Xinyu Zhao, Swarat Chaudhuri, and Greg Durrett.
\newblock Flexible generation of natural language deductions.
\newblock In Marie-Francine Moens, Xuanjing Huang, Lucia Specia, and Scott
  Wen-tau Yih (eds.), \emph{Proceedings of the 2021 Conference on Empirical
  Methods in Natural Language Processing}, pp.\  6266--6278, Online and Punta
  Cana, Dominican Republic, November 2021. Association for Computational
  Linguistics.
\newblock \doi{10.18653/v1/2021.emnlp-main.506}.
\newblock URL \url{https://aclanthology.org/2021.emnlp-main.506/}.

\bibitem[Chen et~al.(2024)Chen, Lin, Sch{\"a}rli, and Zhou]{chen2024teaching}
Xinyun Chen, Maxwell Lin, Nathanael Sch{\"a}rli, and Denny Zhou.
\newblock Teaching large language models to self-debug.
\newblock In \emph{The Twelfth International Conference on Learning
  Representations}, 2024.
\newblock URL \url{https://openreview.net/forum?id=KuPixIqPiq}.

\bibitem[Dalvi et~al.(2021)Dalvi, Jansen, Tafjord, Xie, Smith, Pipatanangkura,
  and Clark]{dalvi-etal-2021-explaining}
Bhavana Dalvi, Peter Jansen, Oyvind Tafjord, Zhengnan Xie, Hannah Smith,
  Leighanna Pipatanangkura, and Peter Clark.
\newblock Explaining answers with entailment trees.
\newblock In Marie-Francine Moens, Xuanjing Huang, Lucia Specia, and Scott
  Wen-tau Yih (eds.), \emph{Proceedings of the 2021 Conference on Empirical
  Methods in Natural Language Processing}, pp.\  7358--7370, Online and Punta
  Cana, Dominican Republic, November 2021. Association for Computational
  Linguistics.
\newblock \doi{10.18653/v1/2021.emnlp-main.585}.
\newblock URL \url{https://aclanthology.org/2021.emnlp-main.585/}.

\bibitem[de~Moura \& Bjørner(2008)de~Moura and Bjørner]{deMoura2008Z3AE}
Leonardo de~Moura and Nikolaj Bjørner.
\newblock {Z3: An Efficient SMT Solver}.
\newblock In \emph{TACAS}, pp.\  337--340, 2008.

\bibitem[{DeepSeek AI}(2025)]{deepseekR1}
{DeepSeek AI}.
\newblock Deepseek-r1.
\newblock \url{https://en.wikipedia.org/wiki/DeepSeek}, 2025.
\newblock Accessed: 2025-08-09.

\bibitem[Guha et~al.(2023)Guha, Nyarko, Ho, Re, Chilton, Narayana,
  Chohlas-Wood, Peters, Waldon, Rockmore, Zambrano, Talisman, Hoque, Surani,
  Fagan, Sarfaty, Dickinson, Porat, Hegland, Wu, Nudell, Niklaus, Nay, Choi,
  Tobia, Hagan, Ma, Livermore, Rasumov-Rahe, Holzenberger, Kolt, Henderson,
  Rehaag, Goel, Gao, Williams, Gandhi, Zur, Iyer, and Li]{guha2023legalbench}
Neel Guha, Julian Nyarko, Daniel~E. Ho, Christopher Re, Adam Chilton, Aditya
  Narayana, Alex Chohlas-Wood, Austin Peters, Brandon Waldon, Daniel Rockmore,
  Diego Zambrano, Dmitry Talisman, Enam Hoque, Faiz Surani, Frank Fagan, Galit
  Sarfaty, Gregory~M. Dickinson, Haggai Porat, Jason Hegland, Jessica Wu, Joe
  Nudell, Joel Niklaus, John~J Nay, Jonathan~H. Choi, Kevin Tobia, Margaret
  Hagan, Megan Ma, Michael Livermore, Nikon Rasumov-Rahe, Nils Holzenberger,
  Noam Kolt, Peter Henderson, Sean Rehaag, Sharad Goel, Shang Gao, Spencer
  Williams, Sunny Gandhi, Tom Zur, Varun Iyer, and Zehua Li.
\newblock Legalbench: A collaboratively built benchmark for measuring legal
  reasoning in large language models.
\newblock In \emph{Thirty-seventh Conference on Neural Information Processing
  Systems Datasets and Benchmarks Track}, 2023.
\newblock URL \url{https://openreview.net/forum?id=WqSPQFxFRC}.

\bibitem[Holzenberger et~al.(2020)Holzenberger, Blair-Stanek, and
  Durme]{holzenberger2020datasetstatutoryreasoningtax}
Nils Holzenberger, Andrew Blair-Stanek, and Benjamin~Van Durme.
\newblock A dataset for statutory reasoning in tax law entailment and question
  answering, 2020.
\newblock URL \url{https://arxiv.org/abs/2005.05257}.

\bibitem[Ji et~al.(2023)Ji, Lee, Frieske, Yu, Su, Xu, Ishii, Bang, Madotto, and
  Fung]{ji-etal-2023-survey}
Ziwei Ji, Nayeon Lee, Rita Frieske, Tiezheng Yu, Dan Su, Yan Xu, Etsuko Ishii,
  Ye~Jin Bang, Andrea Madotto, and Pascale Fung.
\newblock Survey of hallucination in natural language generation.
\newblock \emph{ACM Comput. Surv.}, 55\penalty0 (12), March 2023.
\newblock ISSN 0360-0300.
\newblock \doi{10.1145/3571730}.
\newblock URL \url{https://doi.org/10.1145/3571730}.

\bibitem[Jung et~al.(2022)Jung, Qin, Welleck, Brahman, Bhagavatula, Le~Bras,
  and Choi]{jung-etal-2022-maieutic}
Jaehun Jung, Lianhui Qin, Sean Welleck, Faeze Brahman, Chandra Bhagavatula,
  Ronan Le~Bras, and Yejin Choi.
\newblock Maieutic prompting: Logically consistent reasoning with recursive
  explanations.
\newblock In Yoav Goldberg, Zornitsa Kozareva, and Yue Zhang (eds.),
  \emph{Proceedings of the 2022 Conference on Empirical Methods in Natural
  Language Processing}, pp.\  1266--1279, Abu Dhabi, United Arab Emirates,
  December 2022. Association for Computational Linguistics.
\newblock \doi{10.18653/v1/2022.emnlp-main.82}.
\newblock URL \url{https://aclanthology.org/2022.emnlp-main.82/}.

\bibitem[Lalwani et~al.(2025)Lalwani, Kim, Chopra, Hahn, Jin, and
  Sachan]{lalwani2025auto}
Abhinav Lalwani, Tasha Kim, Lovish Chopra, Christopher Hahn, Zhijing Jin, and
  Mrinmaya Sachan.
\newblock Autoformalizing natural language to first-order logic: A case study
  in logical fallacy detection, 2025.
\newblock URL \url{https://arxiv.org/abs/2405.02318}.

\bibitem[Leang et~al.(2025)Leang, Hong, Li, and Cohen]{leang-etal-2025-theorem}
Joshua Ong~Jun Leang, Giwon Hong, Wenda Li, and Shay~B Cohen.
\newblock Theorem prover as a judge for synthetic data generation.
\newblock In Wanxiang Che, Joyce Nabende, Ekaterina Shutova, and Mohammad~Taher
  Pilehvar (eds.), \emph{Proceedings of the 63rd Annual Meeting of the
  Association for Computational Linguistics (Volume 1: Long Papers)}, pp.\
  29941--29977, Vienna, Austria, July 2025. Association for Computational
  Linguistics.
\newblock ISBN 979-8-89176-251-0.
\newblock \doi{10.18653/v1/2025.acl-long.1448}.
\newblock URL \url{https://aclanthology.org/2025.acl-long.1448/}.

\bibitem[Ling et~al.(2023)Ling, Fang, Li, Huang, Lee, Memisevic, and
  Su]{ling2023deductive}
Zhan Ling, Yunhao Fang, Xuanlin Li, Zhiao Huang, Mingu Lee, Roland Memisevic,
  and Hao Su.
\newblock Deductive verification of chain-of-thought reasoning.
\newblock In \emph{Thirty-seventh Conference on Neural Information Processing
  Systems}, 2023.
\newblock URL \url{https://openreview.net/forum?id=I5rsM4CY2z}.

\bibitem[Nentidis et~al.(2023)Nentidis, Katsimpras, Krithara, Lima~López,
  Farré-Maduell, Gasco, Krallinger, and Paliouras]{Nentidis_2023}
Anastasios Nentidis, Georgios Katsimpras, Anastasia Krithara, Salvador
  Lima~López, Eulália Farré-Maduell, Luis Gasco, Martin Krallinger, and
  Georgios Paliouras.
\newblock \emph{Overview of BioASQ 2023: The Eleventh BioASQ Challenge on
  Large-Scale Biomedical Semantic Indexing and Question Answering}, pp.\
  227–250.
\newblock Springer Nature Switzerland, 2023.
\newblock ISBN 9783031424489.
\newblock \doi{10.1007/978-3-031-42448-9_19}.
\newblock URL \url{http://dx.doi.org/10.1007/978-3-031-42448-9_19}.

\bibitem[Nentidis et~al.(2024)Nentidis, Katsimpras, Krithara, Lima-L{\'o}pez,
  Farr{\'e}-Maduell, Krallinger, Loukachevitch, Davydova, Tutubalina, and
  Paliouras]{450}
Anastasios Nentidis, Georgios Katsimpras, Anastasia Krithara, Salvador
  Lima-L{\'o}pez, Eul{\`a}lia Farr{\'e}-Maduell, Martin Krallinger, Natalia
  Loukachevitch, Vera Davydova, Elena Tutubalina, and Georgios Paliouras.
\newblock Overview of~bioasq 2024: The twelfth bioasq challenge on~large-scale
  biomedical semantic indexing and~question answering.
\newblock In Lorraine Goeuriot, Philippe Mulhem, Georges Qu{\'e}not, Didier
  Schwab, Giorgio~Maria Di~Nunzio, Laure Soulier, Petra Galu{\v s}{\v
  c}{\'a}kov{\'a}, Alba Garc{\'\i}a Seco~de Herrera, Guglielmo Faggioli, and
  Nicola Ferro (eds.), \emph{Experimental IR Meets Multilinguality,
  Multimodality, and Interaction}, Cham, 2024. Springer Nature Switzerland,
  Springer Nature Switzerland.
\newblock ISBN 978-3-031-71908-0.
\newblock \doi{https://doi.org/10.1007/978-3-031-71908-0_1}.
\newblock URL
  \url{https://link.springer.com/chapter/10.1007/978-3-031-71908-0_1}.

\bibitem[Olausson et~al.(2023)Olausson, Gu, Lipkin, Zhang, Solar-Lezama,
  Tenenbaum, and Levy]{olausson-etal-2023-linc}
Theo Olausson, Alex Gu, Ben Lipkin, Cedegao Zhang, Armando Solar-Lezama, Joshua
  Tenenbaum, and Roger Levy.
\newblock {LINC}: A neurosymbolic approach for logical reasoning by combining
  language models with first-order logic provers.
\newblock In Houda Bouamor, Juan Pino, and Kalika Bali (eds.),
  \emph{Proceedings of the 2023 Conference on Empirical Methods in Natural
  Language Processing}, pp.\  5153--5176, Singapore, December 2023. Association
  for Computational Linguistics.
\newblock \doi{10.18653/v1/2023.emnlp-main.313}.
\newblock URL \url{https://aclanthology.org/2023.emnlp-main.313/}.

\bibitem[Olausson et~al.(2024)Olausson, Inala, Wang, Gao, and
  Solar{-}Lezama]{olausson2024is}
Theo~X. Olausson, Jeevana~Priya Inala, Chenglong Wang, Jianfeng Gao, and
  Armando Solar{-}Lezama.
\newblock Is self-repair a silver bullet for code generation?
\newblock In \emph{The Twelfth International Conference on Learning
  Representations, {ICLR} 2024, Vienna, Austria, May 7-11, 2024}, 2024.
\newblock URL \url{https://openreview.net/forum?id=y0GJXRungR}.

\bibitem[{OpenAI}(2024{\natexlab{a}})]{o1reasoningerrors}
{OpenAI}.
\newblock Limitations of reasoning models.
\newblock \url{https://openai.com/index/openai-o1-system-card/},
  2024{\natexlab{a}}.
\newblock Describes common reasoning errors despite correct answers; Accessed:
  2025-08-09.

\bibitem[{OpenAI}(2024{\natexlab{b}})]{o1systemcard}
{OpenAI}.
\newblock Openai o1 system card.
\newblock \url{https://openai.com/index/openai-o1-system-card/},
  2024{\natexlab{b}}.
\newblock Accessed: 2025-08-09.

\bibitem[Pan et~al.(2023)Pan, Albalak, Wang, and Wang]{pan-etal-2023-logic}
Liangming Pan, Alon Albalak, Xinyi Wang, and William Wang.
\newblock Logic-{LM}: Empowering large language models with symbolic solvers
  for faithful logical reasoning.
\newblock In Houda Bouamor, Juan Pino, and Kalika Bali (eds.), \emph{Findings
  of the Association for Computational Linguistics: EMNLP 2023}, pp.\
  3806--3824, Singapore, December 2023. Association for Computational
  Linguistics.
\newblock \doi{10.18653/v1/2023.findings-emnlp.248}.
\newblock URL \url{https://aclanthology.org/2023.findings-emnlp.248/}.

\bibitem[Paul et~al.(2024)Paul, Ismayilzada, Peyrard, Borges, Bosselut, West,
  and Faltings]{paul-etal-2024-refiner}
Debjit Paul, Mete Ismayilzada, Maxime Peyrard, Beatriz Borges, Antoine
  Bosselut, Robert West, and Boi Faltings.
\newblock {REFINER}: Reasoning feedback on intermediate representations.
\newblock In Yvette Graham and Matthew Purver (eds.), \emph{Proceedings of the
  18th Conference of the European Chapter of the Association for Computational
  Linguistics (Volume 1: Long Papers)}, pp.\  1100--1126, St. Julian{'}s,
  Malta, March 2024. Association for Computational Linguistics.
\newblock \doi{10.18653/v1/2024.eacl-long.67}.
\newblock URL \url{https://aclanthology.org/2024.eacl-long.67/}.

\bibitem[Peng et~al.(2023)Peng, Galley, He, Cheng, Xie, Hu, Huang, Liden, Yu,
  Chen, and Gao]{peng2023check}
Baolin Peng, Michel Galley, Pengcheng He, Hao Cheng, Yujia Xie, Yu~Hu, Qiuyuan
  Huang, Lars Liden, Zhou Yu, Weizhu Chen, and Jianfeng Gao.
\newblock Check your facts and try again: Improving large language models with
  external knowledge and automated feedback, 2023.
\newblock URL \url{https://arxiv.org/abs/2302.12813}.

\bibitem[Poesia et~al.(2024)Poesia, Gandhi, Zelikman, and
  Goodman]{poesia2024certified}
Gabriel Poesia, Kanishk Gandhi, Eric Zelikman, and Noah Goodman.
\newblock Certified deductive reasoning with language models.
\newblock \emph{Transactions on Machine Learning Research}, 2024.
\newblock ISSN 2835-8856.
\newblock URL \url{https://openreview.net/forum?id=yXnwrs2Tl6}.

\bibitem[Quan et~al.(2024{\natexlab{a}})Quan, Valentino, Dennis, and
  Freitas]{quan-etal-2024-enhancing}
Xin Quan, Marco Valentino, Louise Dennis, and Andre Freitas.
\newblock Enhancing ethical explanations of large language models through
  iterative symbolic refinement.
\newblock In Yvette Graham and Matthew Purver (eds.), \emph{Proceedings of the
  18th Conference of the European Chapter of the Association for Computational
  Linguistics (Volume 1: Long Papers)}, pp.\  1--22, St. Julian{'}s, Malta,
  March 2024{\natexlab{a}}. Association for Computational Linguistics.
\newblock \doi{10.18653/v1/2024.eacl-long.1}.
\newblock URL \url{https://aclanthology.org/2024.eacl-long.1/}.

\bibitem[Quan et~al.(2024{\natexlab{b}})Quan, Valentino, Dennis, and
  Freitas]{quan-etal-2024-verification}
Xin Quan, Marco Valentino, Louise~A. Dennis, and Andre Freitas.
\newblock Verification and refinement of natural language explanations through
  {LLM}-symbolic theorem proving.
\newblock In Yaser Al-Onaizan, Mohit Bansal, and Yun-Nung Chen (eds.),
  \emph{Proceedings of the 2024 Conference on Empirical Methods in Natural
  Language Processing}, pp.\  2933--2958, Miami, Florida, USA, November
  2024{\natexlab{b}}. Association for Computational Linguistics.
\newblock \doi{10.18653/v1/2024.emnlp-main.172}.
\newblock URL \url{https://aclanthology.org/2024.emnlp-main.172/}.

\bibitem[{Qwen Team}(2024)]{qwen2.5}
{Qwen Team}.
\newblock Qwen2.5: A party of foundation models.
\newblock Blog post; recommended citation for Qwen-2.5, September 2024.
\newblock Includes Qwen2.5-7B and instruction-tuned variants.

\bibitem[Rafailov et~al.(2024)Rafailov, Sharma, Mitchell, Ermon, Manning, and
  Finn]{rafailov2024directpreferenceoptimizationlanguage}
Rafael Rafailov, Archit Sharma, Eric Mitchell, Stefano Ermon, Christopher~D.
  Manning, and Chelsea Finn.
\newblock Direct preference optimization: Your language model is secretly a
  reward model, 2024.
\newblock URL \url{https://arxiv.org/abs/2305.18290}.

\bibitem[Sprague et~al.(2022)Sprague, Bostrom, Chaudhuri, and
  Durrett]{sprague-etal-2022-natural}
Zayne Sprague, Kaj Bostrom, Swarat Chaudhuri, and Greg Durrett.
\newblock Natural language deduction with incomplete information.
\newblock In Yoav Goldberg, Zornitsa Kozareva, and Yue Zhang (eds.),
  \emph{Proceedings of the 2022 Conference on Empirical Methods in Natural
  Language Processing}, pp.\  8230--8258, Abu Dhabi, United Arab Emirates,
  December 2022. Association for Computational Linguistics.
\newblock \doi{10.18653/v1/2022.emnlp-main.564}.
\newblock URL \url{https://aclanthology.org/2022.emnlp-main.564/}.

\bibitem[Tafjord et~al.(2021)Tafjord, Dalvi, and
  Clark]{tafjord-etal-2021-proofwriter}
Oyvind Tafjord, Bhavana Dalvi, and Peter Clark.
\newblock {P}roof{W}riter: Generating implications, proofs, and abductive
  statements over natural language.
\newblock In Chengqing Zong, Fei Xia, Wenjie Li, and Roberto Navigli (eds.),
  \emph{Findings of the Association for Computational Linguistics: ACL-IJCNLP
  2021}, pp.\  3621--3634, Online, August 2021. Association for Computational
  Linguistics.
\newblock \doi{10.18653/v1/2021.findings-acl.317}.
\newblock URL \url{https://aclanthology.org/2021.findings-acl.317/}.

\bibitem[Tafjord et~al.(2022)Tafjord, Dalvi~Mishra, and
  Clark]{tafjord-etal-2022-entailer}
Oyvind Tafjord, Bhavana Dalvi~Mishra, and Peter Clark.
\newblock Entailer: Answering questions with faithful and truthful chains of
  reasoning.
\newblock In Yoav Goldberg, Zornitsa Kozareva, and Yue Zhang (eds.),
  \emph{Proceedings of the 2022 Conference on Empirical Methods in Natural
  Language Processing}, pp.\  2078--2093, Abu Dhabi, United Arab Emirates,
  December 2022. Association for Computational Linguistics.
\newblock \doi{10.18653/v1/2022.emnlp-main.134}.
\newblock URL \url{https://aclanthology.org/2022.emnlp-main.134/}.

\bibitem[Vacareanu et~al.(2024)Vacareanu, Pratik, Spiliopoulou, Qi, Paolini,
  John, Ma, Benajiba, and
  Ballesteros]{vacareanu2024generalpurposeverificationchain}
Robert Vacareanu, Anurag Pratik, Evangelia Spiliopoulou, Zheng Qi, Giovanni
  Paolini, Neha~Anna John, Jie Ma, Yassine Benajiba, and Miguel Ballesteros.
\newblock General purpose verification for chain of thought prompting, 2024.
\newblock URL \url{https://arxiv.org/abs/2405.00204}.

\bibitem[Wei et~al.(2022)Wei, Wang, Schuurmans, Bosma, brian ichter, Xia, Chi,
  Le, and Zhou]{wei2022chain}
Jason Wei, Xuezhi Wang, Dale Schuurmans, Maarten Bosma, brian ichter, Fei Xia,
  Ed~H. Chi, Quoc~V Le, and Denny Zhou.
\newblock Chain of thought prompting elicits reasoning in large language
  models.
\newblock In Alice~H. Oh, Alekh Agarwal, Danielle Belgrave, and Kyunghyun Cho
  (eds.), \emph{Advances in Neural Information Processing Systems}, 2022.
\newblock URL \url{https://openreview.net/forum?id=_VjQlMeSB_J}.

\bibitem[Weir et~al.(2024)Weir, Clark, and Van~Durme]{weir-etal-2024-nellie}
Nathaniel Weir, Peter Clark, and Benjamin Van~Durme.
\newblock Nellie: A neuro-symbolic inference engine for grounded,
  compositional, and explainable reasoning.
\newblock 2024.

\bibitem[Xu et~al.(2024)Xu, Fei, Pan, Liu, Lee, and Hsu]{xu-etal-2024-faithful}
Jundong Xu, Hao Fei, Liangming Pan, Qian Liu, Mong-Li Lee, and Wynne Hsu.
\newblock Faithful logical reasoning via symbolic chain-of-thought.
\newblock In Lun-Wei Ku, Andre Martins, and Vivek Srikumar (eds.),
  \emph{Proceedings of the 62nd Annual Meeting of the Association for
  Computational Linguistics (Volume 1: Long Papers)}, pp.\  13326--13365,
  Bangkok, Thailand, August 2024. Association for Computational Linguistics.
\newblock \doi{10.18653/v1/2024.acl-long.720}.
\newblock URL \url{https://aclanthology.org/2024.acl-long.720/}.

\bibitem[Ye et~al.(2023)Ye, Chen, Dillig, and Durrett]{ye2023satlm}
Xi~Ye, Qiaochu Chen, Isil Dillig, and Greg Durrett.
\newblock Sat{LM}: Satisfiability-aided language models using declarative
  prompting.
\newblock In \emph{Thirty-seventh Conference on Neural Information Processing
  Systems}, 2023.
\newblock URL \url{https://openreview.net/forum?id=TqW5PL1Poi}.

\end{thebibliography}
